\documentclass[10pt,twocolumn,letterpaper]{article}
\usepackage[pagenumbers]{cvpr}
\usepackage{marvosym}
\usepackage{graphicx}
\usepackage{multirow}
\usepackage{float}
\usepackage{subfloat}
\usepackage[linesnumbered,ruled]{algorithm2e}
\usepackage{amsthm,amssymb}
\newtheorem{theorem}{Theorem}
\newtheorem{lemma}[theorem]{Lemma}
\newtheorem{proposition}[theorem]{Proposition}

\definecolor{cvprblue}{rgb}{0.21,0.49,0.74}
\usepackage[pagebackref,breaklinks,colorlinks,allcolors=cvprblue]{hyperref}

\def\paperID{17607}
\def\confName{CVPR}
\def\confYear{2025}

\title{Seeking Consistent Flat Minima for Better Domain Generalization \\via Refining Loss Landscapes}

\author{Aodi Li$^1$, Liansheng Zhuang$^{1}$\textsuperscript{,(\Letter)} Xiao Long$^1$, Minghong Yao$^1$, Shafei Wang$^2$\\
$^1$University of Science and Technology of China, Hefei 230026, China\\
$^2$Peng Cheng Laboratory, Shenzhen 518000, China\\
{\tt\small aodili@mail.ustc.edu.cn, lszhuang@ustc.edu.cn}
}

\begin{document}
\maketitle
\begin{abstract}
Domain generalization aims to learn a model from multiple training domains and generalize it to unseen test domains. 
Recent theory has shown that seeking the deep models, whose parameters lie in the flat minima of the loss landscape, can significantly reduce the out-of-domain generalization error.
However, existing methods often neglect the consistency of loss landscapes in different domains, resulting in models that are not simultaneously in the optimal flat minima in all domains, which limits their generalization ability.
To address this issue, this paper proposes an iterative Self-Feedback Training (SFT) framework to seek consistent flat minima that are shared across different domains by progressively refining loss landscapes during training. 
It alternatively generates a feedback signal by measuring the inconsistency of loss landscapes in different domains and refines these loss landscapes for greater consistency using this feedback signal. 
Benefiting from the consistency of the flat minima within these refined loss landscapes, our SFT helps achieve better out-of-domain generalization.
Extensive experiments on DomainBed demonstrate superior performances of SFT when compared to state-of-the-art sharpness-aware methods and other prevalent DG baselines. 
On average across five DG benchmarks, SFT surpasses the sharpness-aware minimization by 2.6\% with ResNet-50 and 1.5\% with ViT-B/16, respectively.

\end{abstract}

\section{Introduction}
The task of Domain Generalization (DG) is to learn a model from multiple training domains so that it can generalize well to unseen test domains~\cite{DGSurvey}. Though modern deep learning has achieved remarkable success in many areas~\cite{CV,NLP,speech,nlpspeech}, it assumes that training and test data are independent and identically distributed (IID). This assumption is often violated in real applications, which significantly degrades the performance of deep models~\cite{DGSurvey}. 
To address this problem, abundant domain generalization (DG) algorithms have been developed from different perspectives, e.g., invariant or causal representation learning~\cite{DANN,IRM,stable}, disentangled representation learning~\cite{Sagnet,DecAug}, distributionally robust optimization~\cite{DRO}, meta-learning~\cite{MLDG, MetaReg, MetaVIB}, data augmentation~\cite{Mixup, ADA}, etc. 
Nonetheless, it is still an open problem for the communities of modern machine learning. 

\begin{figure}[t]
\centering
\subfloat[Inconsistent loss landscapes]{\label{fig:main}
\includegraphics[scale=0.35]{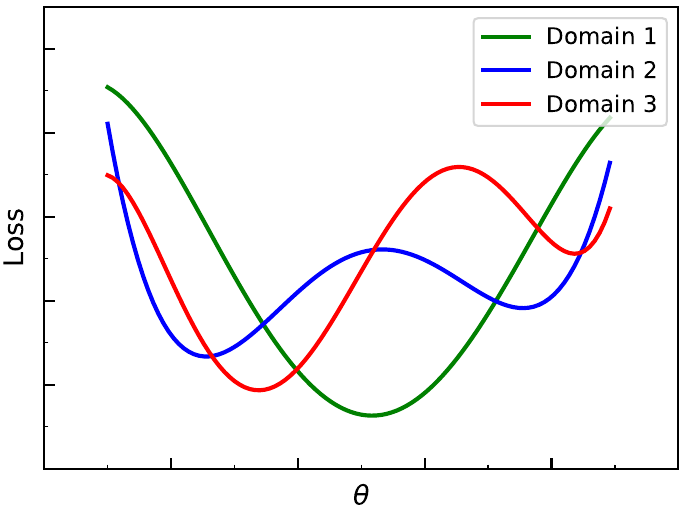}}
\hspace{-0.2cm}
\subfloat[Loss landscapes with consistency]{
\includegraphics[scale=0.35]{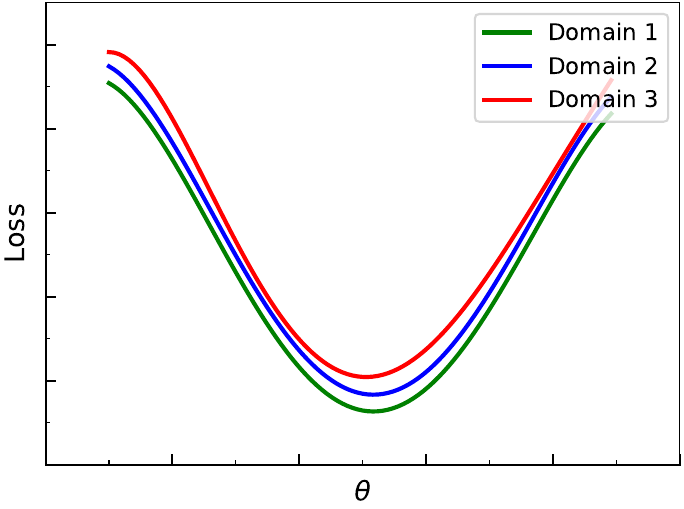}}
\vspace{-0.5em}
\caption{\textbf{Loss Landscapes without and with consistency.} The left subplot (a) illustrates the inconsistency of loss landscapes across different domains, which arise due to domain shifts. This paper proposes refining these landscapes to achieve improved consistency, as demonstrated in the right subplot (b).}
\vspace{-1.5em}
\end{figure}

Recently, the loss landscape~\cite{LossVisual}, which refers to the shape of a loss function within the parameter space, provides a unified perspective for understanding various DG algorithms.  
Essentially, the training process of deep models is to seek the minimum point within the training loss landscape.
From the view of loss landscapes~\cite{LossVisual}, the ultimate goal of various DG methods is to construct a suitable loss landscape so that the loss minimum point sought on training domains is exactly the loss minimum point on unseen test domains.
In fact, significant studies have revealed the connection between the geometry of the loss landscape (particularly the flatness of minima~\cite{Sharpness1994}) and generalization from both theoretical and empirical perspectives~\cite{SharpGenelink,SharpGenelink2,SharpGenelink3,SAM}. 
This connection has shown potential in enabling novel methods to model training that yield better generalization~\cite{SAM}.
For example, by penalizing the sharpness of the loss landscape, numerous methods~\cite{SAM,sharpdiffusion,entropySGD} have exhibited superior generalization on the DomainBed benchmark~\cite{DomainBed} and many other datasets~\cite{NLPSAM,SpeechSAM}. 
However, in the DG scenario, the domain shifts contribute to the drastic discrepancy of loss landscapes~\cite{SWAD}, and thus loss landscapes in test domains may not exhibit similar geometries as in training domains. 
Consequently, the flat minima sought on the training domains do not have a low loss value in the unseen test domains, meaning that the learned models may have inferior performance in the test domains. 
Therefore, maintaining consistency of the loss landscapes between the training and unseen test domains is crucial for better generalization, as it enables us to seek the consistent flat minima that perform well across all domains.

To address this issue, this paper proposes refining the loss landscapes to maintain their consistency in different domains. However, there are several primary challenges to be addressed. 
Firstly, in the DG setting, the inaccessibility of the test domain restricts us to using limited training domains for refining loss landscapes. Then, how to ensure that 
the loss landscape consistency achieved on training domains can transfer to unseen test domains
becomes a challenging endeavor.  
Secondly, refining the loss landscapes usually means high complexity. Typically, it involves the selection of different model architectures~\cite{Nas-ood, ViTOoD, ViTOoDTwo, AdaptNAS} or the artificial design of novel loss functions~\cite{IRM,VREx,ARM}. Architecture selection suffers from substantial computational overhead~\cite{NAS1000}, while artificially designing an effective loss function for DG is also exceedingly intractable.
Therefore, finding efficient approaches to refine various training loss landscapes is far from straightforward.

Inspired by studies on the self-refinement of large-scale language models~\cite{selfrefineLLM,selfLLM}, this paper introduces an iterative two-phase framework called \emph{Self-Feedback Training} (SFT) to progressively refine loss landscapes and find consistent flat minima across different domains. 
Specifically, it alternatively generates a feedback signal by measuring the inconsistency of loss landscapes in different domains in the \emph{feedback} phase, and then refines these loss landscapes for greater consistency using this feedback signal in the \emph{refinement} phase.
To ensure the consistency of loss landscapes between the training and unseen test domains, this paper introduces a training-domain split scheme in the feedback phase. 
In each iteration, the SFT quantifies the inconsistency by comparing the loss landscapes of the trained and held-out domains.
Theoretical analysis (in the supplementary file) shows that in this way, the sharpness of the test loss can be constrained by that of the training domains, which also means the transferability of the loss landscapes consistency. 
Furthermore, to enable low-complexity refinement of loss landscapes, a landscape refiner is introduced in the refinement phase. Instead of altering the model architecture or devising novel loss functions, 
the refiner continuously modifies the geometries of the loss landscapes by generating dynamical soft labels as training progresses.
Compared to existing methods~\cite{Nas-ood,AdaptNAS}, our landscape refining method not only substantially reduces the computational overhead, but also overcomes the over-confident issue brought by traditional one-hot labels~\cite{LabelSmooth}.
To our best knowledge, this is the first work to refine the loss landscapes from a label-based perspective. 
Benefiting from the consistency of the flat minima within these refined loss landscapes, our SFT helps achieve better out-of-domain generalization.
Extensive experiments on a synthetic dataset and five popular DG benchmarks demonstrate that SFT achieves superior performance compared to popular sharpness-aware methods~\cite{SAM,GAM,GSAM,FAD,SAGM} and other prevalent DG methods.

In summary, our key contributions are as follows:
\begin{itemize}[leftmargin=15pt]

\item A novel self-feedback training framework, which could refine loss landscapes and find consistent flat minima across different domains, is proposed to obtain models with better domain generalization. 

\item
A landscape refiner is introduced to facilitate the efficient refinement of loss landscapes. 
To our best knowledge, it is the first work to refine the loss geometry from the perspective of labels.

\item 
 Extensive experiments on the synthetic dataset and five popular DG benchmarks demonstrate the superiority of SFT over popular sharpness-aware methods and other prevalent DG methods. 

\end{itemize}

\section{Related Work}
In this section, we review related areas of research: studies on loss landscapes for model generalization and domain generalization.

\textbf{Loss landscapes for model generalization.} 
The idea of searching for ``flat'' minima of loss landscapes can be traced back to Hochreiter and Schmidhuber~\cite{flat1997}, who proposed penalizing the sharpness of minima as a form of regularization, related to the Minimum Description Length (MDL) principle. Over the past few decades, numerous studies~\cite{SharpGenelink,SharpGenelink2,SharpGenelink3} have linked the flatness of loss landscapes to the generalization ability of deep models. As a result, many optimization methods have been developed to find flatter minima, such as Entropy-SGD~\cite{entropySGD} and the diffusion approach~\cite{sharpdiffusion}. More recently, two methods, Stochastic Weight Averaging (SWA)~\cite{SWA} and Sharpness-Aware Minimization (SAM)~\cite{SAM}, have gained significant attention for their effectiveness and scalability.
SWA finds flatter minima by averaging model parameters across the optimization trajectory, while SAM seeks parameters within regions of the loss landscape that exhibit uniformly low loss values. Jean Kaddour \emph{et al.} ~\cite{SWA_SAM} provided a detailed comparison of these methods, showing that SAM tends to lead models to wider basins, while SWA merely helps locate the center of one certain basin. 
Several efficient variants of SAM~\cite{SAF,AESAM,SSAM} have also been developed for real-world applications.
In this paper, we focus on addressing the inconsistency of loss landscapes when SAM is applied in DG scenarios.

\textbf{Loss landscapes for domain generalization.} 
Recently, optimization techniques aiming at finding flat minima in loss landscapes have also been applied to improve the out-of-domain generalization of deep models~\cite{SAMDG2024}.
Junbum Cha \emph{et al.}~\cite{SWAD} introduced the flatness to the domain generalization and proposed a novel Stochastic Weight Averaging Densely (SWAD), which is an extension of SWA with a dense and overfit-aware stochastic weight sampling strategy. 
Zhang \emph{et al.}~\cite{GAM} pointed out that SAM-based methods are limited to zeroth-order flatness, which may be insufficient to discriminate minima with low generalization error. Thus, they developed a novel method named GAM, which targets first-order flatness to enhance generalization. Moreover, FAD~\cite{FAD} optimizes both zeroth- and first-order flatness simultaneously for domain generalization.
However, these methods primarily focus on seeking better flat minima for DG, ignoring the influence of landscape discrepancy across multiple domains.
In contrast, Giambattista Parascandolo \emph{et al.}~\cite{AndMask} considered the domain discrepancy via formalizing a notion of consistency for minima of loss surfaces, and then proposed an algorithm (ANDMask) to alter the optimization trajectory through masking inconsistent gradient components among training domains. 
Different from all of the above methods that consider loss landscapes to be static, we attempt to refine the loss surfaces dynamically for consistency and finally seek consistent flat minima across various domains.

\section{Methodology}
\subsection{Problem Formulation and Preliminaries}
Let us begin with a formal description of \emph{domain} and \emph{domain generalization (DG)}.
Let $\mathcal{X}$ and $\mathcal{Y}$ denote the input sample space and the category space, respectively. 
In the DG problem, we use data sampled from $p$ training distributions $\{\mathcal{D}_d\}_{d=1}^p$
and $q$ test distributions
$\{\mathcal{D}_d\}_{d=p+1}^{p+q}$,
each of which defines on the joint space $\mathcal{X} \times \mathcal{Y}$.
$D_d =\{(\boldsymbol{x}_i^{(d)},y_i^{(d)})\}_{i=1}^{n_d}$ denotes the dataset sampled from the $d$-th distribution $\mathcal{D}_d$, which is referred to as the $d$-th \emph{domain}. 
$(\boldsymbol{x}_i^{(d)},y_i^{(d)}) \in \mathcal{X} \times \mathcal{Y}$ denotes the $i$-th sample from the domain $\mathcal{D}_d$ and $n_d$ denotes the number of samples in the $d$-th domain. 
Notably, each domain contains different domain statistics but shares the same category space. 
\emph{Domain generalization} aims to train a model $\boldsymbol{f}_{\boldsymbol{\theta}}:\mathcal{X}\rightarrow \mathcal{Y}$ on the $p$ training domains so that it can  generalize well to the $q$ novel target domains. 
Please note that the $q$ test domains are inaccessible during training, which differs from the problem of domain adaptation~\cite{DA}.

In this paper, we consider the model mentioned above to be a parametric deep neural network $\boldsymbol{f}_{\boldsymbol{\theta}}$, with $\boldsymbol{\theta}$ denoting the parameters of the network. 
Standard ERM training of networks usually utilizes the cross entropy (CE) as the loss function:
\begin{equation}
    \mathcal{L}_{D_{tr}}^{\mathrm{CE}}(\boldsymbol{\theta}) = -\sum_{d=1}^{p}\sum_{i=1}^{n_d} \mathbf{y}_i^{(d)\mathbf{T}} \log \boldsymbol{f}_{\boldsymbol{\theta}}(\boldsymbol{x}_i^{(d)}),  
\end{equation}
where $\mathcal{L}_{D_{tr}}^{\mathrm{CE}}$ represents the cross-entropy loss computed on the training set, which is an aggregation of samples from all the training domains. In this paper, we use the superscript on $\mathcal{L}$ to indicate the type of loss function (e.g., cross entropy), while the subscript denotes the dataset on which the loss is calculated.
$\boldsymbol{f}_{\boldsymbol{\theta}}(\boldsymbol{x}_i^{(d)})$ is a vector with $N$ dimensions (where $N$ is the number of classes), and the $j$-th entry of the vector represents the predictive probability that $\boldsymbol{x}_i^{(d)}$ belongs to the $j$-th class. 
$\mathbf{y}_i^{(d)}$ denotes the one-hot encoded label of $y_i^{(d)}$, whose superscript ``$\mathbf{T}$'' indicates the matrix transpose operation.

However, optimizing the training loss value only (e.g., ERM) can easily lead to suboptimal model quality~\cite{SAM}. 
Thus, SAM~\cite{SAM} tries to simultaneously minimize loss value and loss sharpness. 
It defines the sharpness by measuring how quickly the training loss can be increased by adding a small perturbation $\boldsymbol{\epsilon}$ (with norm less than $\rho$) to the current parameter $\boldsymbol{\theta}$. To avoid sharpness being dependent on the perturbation $\boldsymbol{\epsilon}$, the worst case is considered when evaluating sharpness:
\begin{equation}
    \mathcal{L}_{D_{tr}}^{\mathrm{CES}}(\boldsymbol{\theta}) = \max_{|| \boldsymbol{\epsilon} || \leq \rho} \mathcal{L}_{D_{tr}}^{\mathrm{CE}}(\boldsymbol{\theta} + \boldsymbol{\epsilon}) - \mathcal{L}_{D_{tr}}^{\mathrm{CE}}(\boldsymbol{\theta}). 
\end{equation}
Here, we use $\mathcal{L}_{D_{tr}}^{\mathrm{CES}}$ to denote the sharpness measure for $\mathcal{L}_{D_{tr}}^{\mathrm{CE}}$. Since the perturbation strength $\rho$ is small enough, the above optimization could be solved approximately via Taylor expansion, leading to the optimal perturbation $\bar{\boldsymbol{\epsilon}}_{D_{tr}}^{\mathrm{CE}}$ for
$\mathcal{L}_{D_{tr}}^{\mathrm{CE}}$ as:
\begin{equation}
   \bar{\boldsymbol{\epsilon}}_{D_{tr}}^{\mathrm{CE}} (\boldsymbol{\theta}) = \rho \frac{\nabla_{\boldsymbol{\theta}} \mathcal{L}_{D_{tr}}^{\mathrm{CE}}(\boldsymbol{\theta})}{||\nabla_{\boldsymbol{\theta}} \mathcal{L}_{D_{tr}}^{\mathrm{CE}}(\boldsymbol{\theta})||}.
\end{equation}
Then, the final loss function used in SAM combines both the loss value and the loss sharpness, and is given by:
\begin{equation}
    \mathcal{L}_{D_{tr}}^{\mathrm{SAM}}(\boldsymbol{\theta}) =  \mathcal{L}_{D_{tr}}^{\mathrm{CE}}(\boldsymbol{\theta} + \bar{\boldsymbol{\epsilon}}_{D_{tr}}^{\mathrm{CE}}). 
\end{equation}
During training with (stochastic) gradient descent, the contribution of $\nabla_{\boldsymbol{\theta}}\bar{\boldsymbol{\epsilon}}_{D_{tr}}^{\mathrm{CE}} (\boldsymbol{\theta})$ could be neglected due to small perturbation strength $\rho$.
\setlength{\textfloatsep}{5pt}
\begin{algorithm}[t]
\small
\caption{SFT for Domain Generalization.}
\label{alg:SFT}

\KwIn{Training domains $D_{tr}$, learning rate $\eta$ and other hyperparameters.}
\KwOut{Model parameter $\boldsymbol{\theta}$.}

\textbf{Init:} Parameters of the model $\boldsymbol{f}_{\boldsymbol{\theta}}$ and refiner $\boldsymbol{g}_{\boldsymbol{\phi}}$;

\While{not converge}{

    \textbf{The Feedback Phase:}
    
    Randomly select two training domains $D_d$ and $D_{d'}$.

    Update the model parameter $\boldsymbol{\theta}$ on the domain $D_d$:
    $\boldsymbol{\theta} \leftarrow \boldsymbol{\theta} -\eta \nabla_{\boldsymbol{\theta}} \mathcal{L}_{D_{d}}^{\mathrm{SAM-SL}}(\boldsymbol{\theta},\boldsymbol{\phi})$.

    Calculate the feedback signal $|\mathcal{L}_{D_{d}}^{\mathrm{CES-SL}}(\boldsymbol{\theta},\boldsymbol{\phi}) - \mathcal{L}_{D_{d'}}^{\mathrm{CES-SL}}(\boldsymbol{\theta},\boldsymbol{\phi})|$ using both domains.
    
    \textbf{The Refinement Phase:}

    Calculate the total loss for refinement, $\mathcal{L}_{D_{tr}}^{\mathrm{Refine}}(\boldsymbol{\theta},\boldsymbol{\phi})$, by leveraging the above feedback signal.
    
    Update the refiner parameter:
    $\boldsymbol{\phi} \leftarrow \boldsymbol{\phi} -\eta \nabla_{\boldsymbol{\phi}} \mathcal{L}_{D_{tr}}^{\mathrm{Refine}}(\boldsymbol{\theta},\boldsymbol{\phi})$.
    
}
\end{algorithm}

\subsection{Self-Feedback Training}

To address the issue of landscape inconsistency when applying SAM-based methods to DG scenarios, we introduce an iterative two-phase framework called \emph{Self-Feedback Training} (SFT) to progressively refine loss landscapes and find consistent flat minima across different domains.
Specifically, it alternatively generates a feedback signal by measuring the inconsistency of loss landscapes in different domains in the \emph{feedback} phase, and then refines these loss landscapes for greater consistency using this feedback signal in the \emph{refinement} phase.
In the following, this paper provides a detailed description of this framework.

\subsubsection{The Feedback Phase}
In this phase, SFT mainly detects inconsistency between various loss landscapes and generates a feedback signal.

Let's start by analyzing the key elements necessary for obtaining a feedback signal. First, due to the absence of test domains, we are limited to assessing the landscape inconsistency within the training domains. 
Making full use of the training domains to simulate domain shifts that may occur in unseen domains will help improve the model's generalization ability to those unknown domains.
Second, the training loss should depend on some learnable parameters, which enables us to optimize them for landscape refinement in the next phase. Finally, we need to determine which measurement to use when assessing the landscape inconsistency.

Considering the above analysis, we choose to implement a training-domain split scheme, where at each iteration, two training domains, $D_d$ and $D_{d'}$, are randomly selected. $D_d$ is used to train the model, while $D_{d'}$ is held out to evaluate landscape inconsistency in comparison to $D_d$.
During training on the domain $D_d$, we employ the training loss that depends on the parameters of a landscape refiner $\boldsymbol{g}_{\boldsymbol{\phi}}$. This refiner enables the efficient refinement of loss landscapes via generating soft labels for each sample:
\begin{equation}
    \Tilde{\mathbf{y}}_i^{(d)} = \boldsymbol{g}_{\boldsymbol{\phi}}(\boldsymbol{x}_i^{(d)}).
\end{equation}
Then, the loss function for sharpness-aware minimization can be expressed as:
\begin{equation}
\mathcal{L}_{D_{d}}^{\mathrm{SAM-SL}}(\boldsymbol{\theta},\boldsymbol{\phi}) =  \mathcal{L}_{D_{d}}^{\mathrm{CE-SL}}(\boldsymbol{\theta} + \bar{\boldsymbol{\epsilon}}_{D_{d}}^{\mathrm{CE-SL}}, \boldsymbol{\phi}).
\end{equation}
Here, ``SL'' represents the utilization of soft labels. $\mathcal{L}_{D_{d}}^{\mathrm{CE-SL}}$ and $\bar{\boldsymbol{\epsilon}}_{D_{d}}^{\mathrm{CE-SL}}$ are defined as:
\begin{align}
     & \mathcal{L}_{D_{d}}^{\mathrm{CE-SL}}(\boldsymbol{\theta},\boldsymbol{\phi}) = -\sum_{i=1}^{n_d} \Tilde{\mathbf{y}}_i^{(d)\mathbf{T}} \log \boldsymbol{f}_{\boldsymbol{\theta}}(\boldsymbol{x}_i^{(d)}), \\
     & \bar{\boldsymbol{\epsilon}}_{D_{d}}^{\mathrm{CE-SL}} (\boldsymbol{\theta},\boldsymbol{\phi}) = \rho \frac{\nabla_{\boldsymbol{\theta}} \mathcal{L}_{D_{d}}^{\mathrm{CE-SL}}(\boldsymbol{\theta},\boldsymbol{\phi})}{||\nabla_{\boldsymbol{\theta}} \mathcal{L}_{D_{d}}^{\mathrm{CE-SL}}(\boldsymbol{\theta},\boldsymbol{\phi})||}.
\end{align}
For evaluating landscape inconsistency, one direct approach is to look at the difference in loss values across domains. However, the loss value only captures zero-order information. Considering that loss sharpness is closely related to the Hessian matrix and reflects second-order loss information, we instead use the difference in loss sharpness, $|\mathcal{L}_{D_{d}}^{\mathrm{CES-SL}}(\boldsymbol{\theta},\boldsymbol{\phi}) - \mathcal{L}_{D_{d'}}^{\mathrm{CES-SL}}(\boldsymbol{\theta},\boldsymbol{\phi})|$, to quantify the landscape consistency, which serves as a feedback signal to the landscape refiner. Here, $\mathcal{L}_{D_{d}}^{\mathrm{CES-SL}}$ is define as:
\begin{equation}
    \mathcal{L}_{D_{d}}^{\mathrm{CES-SL}}(\boldsymbol{\theta},\boldsymbol{\phi}) =  \mathcal{L}_{D_{d}}^{\mathrm{CE-SL}}(\boldsymbol{\theta} + \bar{\boldsymbol{\epsilon}}_{D_{d}}^{\mathrm{CE-SL}}, \boldsymbol{\phi}) - \mathcal{L}_{D_{d}}^{\mathrm{CE-SL}}(\boldsymbol{\theta},\boldsymbol{\phi}). 
\end{equation}
Similarly, $\mathcal{L}_{D_{d'}}^{\mathrm{CES-SL}}$ can be defined.
Please note that the landscape refiner parameter $\boldsymbol{\phi}$ will be optimized in the refinement phase using the feedback signal and could be treated as constant in this phase.
\begin{figure}[t]
\centering

\subfloat[Domain 1]{\label{fig:loss_visualization_1}

{\includegraphics[scale=0.35]{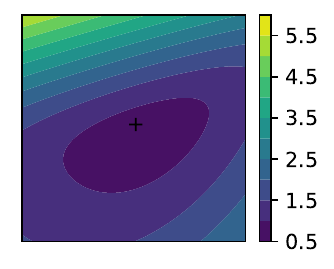}}\vspace{-0.1cm}}
\subfloat[Domain 2]{\label{fig:loss_visualization_2}
   
{\includegraphics[scale=0.35]{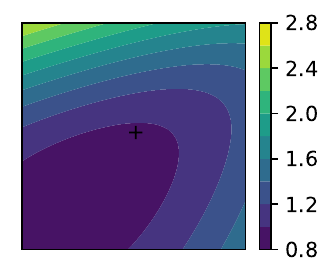}}\vspace{-0.1cm}}
\subfloat[Domain 3]{\label{fig:loss_visualization_3}
   
{\includegraphics[scale=0.35]{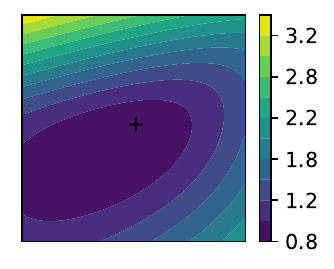}}\vspace{-0.1cm}}
\subfloat[Domain 4]{\label{fig:loss_visualization_4}
   
{\includegraphics[scale=0.35]{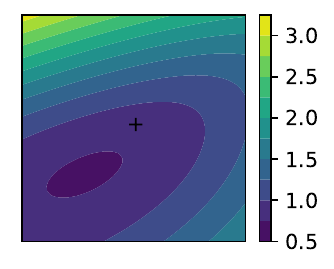}}\vspace{-0.1cm}} \\
\vspace{-0.2em}
\subfloat[Domain 1]{\label{fig:loss_visualization_5}
   
{\includegraphics[scale=0.35]{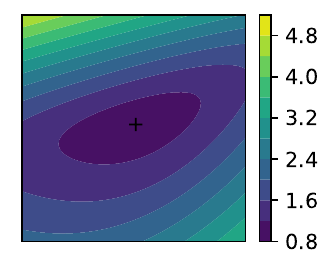}}\vspace{-0.1cm}}
\subfloat[Domain 2]{\label{fig:loss_visualization_6}
   
{\includegraphics[scale=0.35]{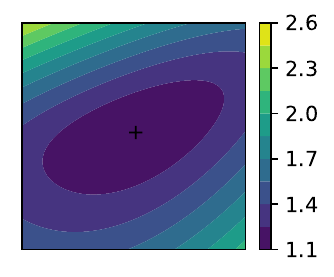}}\vspace{-0.1cm}}
\subfloat[Domain 3]{\label{fig:loss_visualization_7}
   
{\includegraphics[scale=0.35]{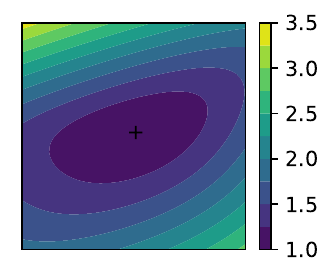}}\vspace{-0.1cm}}
\subfloat[Domain 4]{\label{fig:loss_visualization_8}
   
{\includegraphics[scale=0.35]{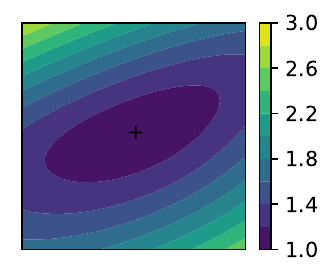}}\vspace{-0.1cm}}
\caption{\textbf{2D visualization of loss surfaces at each domain with/without landscape refinement.} The first row shows the inconsistency of loss surfaces using one-hot labels (without refinement); the second row shows the improved landscape consistency using soft labels generated by the landscape refiner. The final well-trained model is marked by ``$+$''.}\label{fig:loss_visualization}
\end{figure}

\subsubsection{The Refinement Phase}
After obtaining the feedback signal, this phase will use it to optimize the landscape refiner to enhance the consistency of different landscapes. 

As mentioned earlier, the landscape refiner refines the loss landscapes by generating soft labels for subsequent training. There are several important considerations about these labels.
First, it is essential to maintain the correctness of the labels as much as possible, as this directly impacts the effectiveness of training. Second, the soft labels help improve the consistency of the refined loss landscapes, which is central to addressing the landscape inconsistency problem discussed in this paper. Moreover, there are two potential directions to refining the loss landscapes for consistency: making them both sharper or making them both flatter. Given the relationship between flatness and generalization, the soft labels need to be designed to encourage the loss landscapes to become flatter.

To ensure label correctness, we could use the standard cross-entropy loss to force the generated soft labels to approximate one-hot labels. However, this approach carries the risk of overfitting to the one-hot labels, leading to overconfidence and limiting the model's generalization ability. One technique that attempts to mitigate this issue is label smoothing~\cite{LabelSmooth}, but it merely introduces noise to the one-hot labels, making them less informative and unable to capture the underlying relationships between categories and different input samples.
This paper focuses on a more fundamental problem: minimizing the ``distance'' between the refiner's output and the true label space for each category. For instance, if the first category is the true label for the sample $\boldsymbol{x}_i$ (i.e., $y_i=1$), the corresponding label space is defined as:
\begin{equation}
C_1 = \{(q_1,\ldots,q_N) \mid \forall k \neq 1: q_{1} \geq \alpha q_k, \sum_{k=1}^{N} q_k = 1\},
\end{equation}
where $\alpha \geq 1$ is a hyperparameter that represents the minimum ratio between $q_1$ and $q_k$.
Similar to the concept of ``distance'' in Euclidean space, we define the ``distance'' between the output $\Tilde{\mathbf{y}}=\boldsymbol{g}_{\boldsymbol{\phi}}(\boldsymbol{x})$ and the label space $C_1$ as:
\begin{equation}
\min_{\mathbf{y}} \, \mathrm{KL}(\mathbf{y} \, || \, \Tilde{\mathbf{y}}) \quad \text{subject to} \quad \mathbf{y} \in C_1,
\end{equation}
where $\mathrm{KL}(\cdot \| \cdot) $ denotes the Kullback-Leibler (KL) divergence.
We propose an efficient algorithm (Algorithm \ref{alg:PCE}) to find the optimal solution $\mathbf{y}^\star$ for this problem, which converges in $N$ steps ($N$ is the number of classes) and is computationally more efficient than using common convex programming tools. 
For the condition that $y_i\neq 1$, we can also get the optimal solution via category swapping before and after applying Algorithm \ref{alg:PCE}.
Once having $\mathbf{y}^\star$, the task becomes minimizing the KL divergence between $\mathbf{y}^\star$ and $\Tilde{\mathbf{y}}$:  
$\mathrm{KL}(\mathbf{y}^\star \, || \, \Tilde{\mathbf{y}})$.
Since $\mathbf{y}^\star$ depends on $\boldsymbol{\phi}$ and Algorithm \ref{alg:PCE} may involve non-differentiable operations, we adopt a widely-used alternative iterative strategy: we treat $\mathbf{y}^\star$ as constant during each update step for $\boldsymbol{\phi}$, and after updating $\boldsymbol{\phi}$, we recalculate $\mathbf{y}^\star$ using Algorithm \ref{alg:PCE}.
Minimizing the KL divergence then leads to an objective similar to cross entropy:
\begin{equation}
    \mathcal{L}_{D_{tr}}^{\mathrm{PCE}}(\boldsymbol{\phi}) = -\sum_{d=1}^{p}\sum_{i=1}^{n_d} \mathbf{y}_i^{\star(d)\mathbf{T}} \log \boldsymbol{g}_{\boldsymbol{\phi}}(\boldsymbol{x}_i^{(d)}),
\end{equation}
which we refer to as projection cross-entropy (PCE) loss. Interestingly, we find that label smoothing is just a special case of our method, applicable only to a small subset of samples. More detailed derivations and further discussion are available in the \textbf{supplementary file}.

\setlength{\textfloatsep}{5pt}
\begin{algorithm}[t]
\small
\DontPrintSemicolon
\SetAlgoLined
\KwIn {The hyperparameter $\alpha$, $\Tilde{\mathbf{y}}=(p_1, \dots, p_N)$.}
\KwOut {The optimal solution $\mathbf{y}^\star=(q_1, \dots, q_N)$.}
  
  \textbf{Initialization:} $A \gets \emptyset$, $B \gets \{j | \alpha p_j > p_1\}$, $t \gets 1$.\;
  
  Sort the elements of $B$ in descending order: $p_{j_1} \geq \dots \geq p_{j_{|B|}}$.\;
  
  Update: $A \gets A \cup \{j_1\}$, $t \gets t + 1$.\;
  
  \While{$t \leq |B|$}{
    \eIf{$\left(p_1^{\alpha} \left( \prod_{j \in A} \alpha p_j \right) \right)^{\frac{1}{|A| + \alpha}} < \alpha p_{j_t}$}{
      Update: $A \gets A \cup \{j_t\}$, $t \gets t + 1$.\;
    }{
    Break
    }
  }
  
  Calculate $q_1$: $q_1 = \left( p_1^{\alpha} \left( \prod_{j \in A} \alpha p_j \right) \right)^{\frac{1}{|A| + \alpha}}$.\;
  
  \For{$i \in \{2, \dots, N\}$}{
    \eIf{$i \in A$}{
      $q_i = q_1 / \alpha$.
    }{
      $q_i = p_i$.
    }
  }
  
  Normalize $q_1, \dots, q_N$ such that $\sum_i q_i = 1$.\;
  \caption{KL Divergence Minimization}\label{alg:PCE}
\end{algorithm}

To encourage the loss landscapes to become flatter while pursuing consistency, we add the sharpness term to the previously obtained feedback signal (the sharpness difference) as an extra penalty. Then, the final loss function is defined as:
\begin{align}\label{eq:refine}
\mathcal{L}_{D_{tr}}^{\mathrm{Refine}}(\boldsymbol{\theta},\boldsymbol{\phi})
 = \mathcal{L}^{\mathrm{PCE}}_{D_{tr}}(\boldsymbol{\phi}) 
+ \lambda_{1} \mathcal{L}_{D_{d}}^{\mathrm{CES-SL}}(\boldsymbol{\theta},\boldsymbol{\phi}) \nonumber \\
+ \lambda_{2}|\mathcal{L}_{D_{d}}^{\mathrm{CES-SL}}(\boldsymbol{\theta},\boldsymbol{\phi}) - \mathcal{L}_{D_{d'}}^{\mathrm{CES-SL}}(\boldsymbol{\theta},\boldsymbol{\phi})|.
\end{align} 
Here, the hyperparameters $\lambda_{1}$ and $\lambda_{2}$ control the relative importance of each term in the loss function.
By using this loss function to update the landscape refiner $\boldsymbol{g}_{\boldsymbol{\phi}}$, the generated soft labels could help adjust the loss landscapes, making them more consistent and flatter over time. 

As the iterative process of feedback and refinement continues, the model benefits from these refined landscapes, guiding it toward more stable and consistent optimization paths.
Ultimately, the model can seek the consistent flat minima across domains, which helps achieve better out-of-domain generalization.

\section{Experiments}
In this section, we begin by designing a toy experiment to visually demonstrate the landscape consistency achieved through self-feedback training, and explore how the landscape consistency impacts DG performance. A thorough theoretical analysis is provided in the \textbf{supplementary file}. After that, we perform extensive experiments on real-world datasets to validate the effectiveness of the SFT method.

\subsection{Experiments on Toy Dataset}

\textbf{Data generation.}
In our toy experiments, we use a hierarchical Gaussian model. Specifically, we assume that the data from the $i$-th class in the $j$-th domain follows a two-dimensional Gaussian distribution, i.e., 
$
p(\boldsymbol{x}) = \mathcal{N}(\boldsymbol{x}; \boldsymbol{\mu}_{ij}, \boldsymbol{\Sigma}_{ij}),
$
where $\boldsymbol{\mu}_{ij} \in \mathbb{R}^2$ and $\boldsymbol{\Sigma}_{ij} \in \mathbb{R}^{2 \times 2}$ are the mean vector and covariance matrix for the $j$-th domain in the $i$-th class, respectively. The mean vector $\boldsymbol{\mu}_{ij}$ is sampled from another Gaussian distribution:
$
p(\boldsymbol{\mu}_{ij}) = \mathcal{N}(\boldsymbol{\mu}_{ij}; \boldsymbol{\mu}_i, \boldsymbol{\Sigma}_i),
$
where $\boldsymbol{\mu}_i$ and $\boldsymbol{\Sigma}_i$ represent the mean and covariance for the $i$-th class across domains.
In our setup, we generate a dataset with three classes and four domains. The first three domains are used for training, while the fourth domain serves as the test set. Further details can be found in the \textbf{supplementary file}. A linear classifier is employed in these toy experiments.

\textbf{Loss landscape visualization.}
To investigate the consistency of the landscape across the four domains after self-feedback training, we visualize the 2D loss surfaces in Fig.\ref{fig:loss_visualization}.
We first choose three model weights $(\theta_1, \theta_2, \theta_3)$,\footnote{
Here, $\theta_1$ represents the parameters of the well-trained model, while $\theta_2$ and $\theta_3$ correspond to two randomly initialized models.
} and use them to derive two axes, $e_1$ and $e_2$, through the Gram–Schmidt process.
Then, we compute loss values by varying the coefficients $\beta_1$ and $\beta_2$ in the linear combination $\theta_1 + \beta_1 e_1 + \beta_2 e_2$, where $\beta_1,\beta_2 \in [-2,2]$ in our experiments.
From Fig.\ref{fig:loss_visualization}, we observe that the loss surfaces using soft labels generated by the landscape refiner exhibit greater consistency across domains compared to those obtained using one-hot labels. Besides, we find that the final model sought by SFT, marked by a ``+'', indeed locates the flat minima within the more consistent loss surfaces. These results demonstrate the effectiveness of SFT.

\begin{figure}[h]
\centering
\subfloat[Training Domains]{\label{fig:lambda-train}
\includegraphics[scale=0.3]{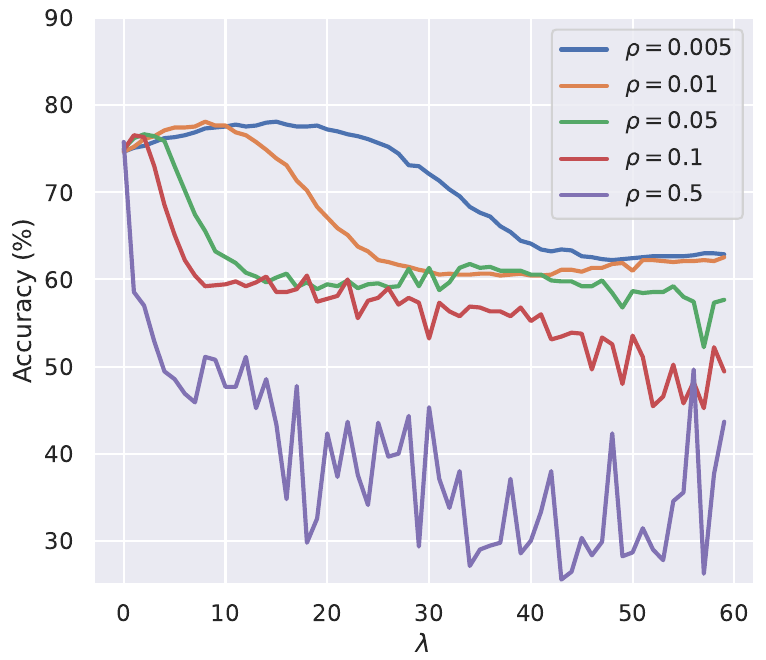}}
\subfloat[Test Domain]{
\label{fig:lambda-test}
\includegraphics[scale=0.3]{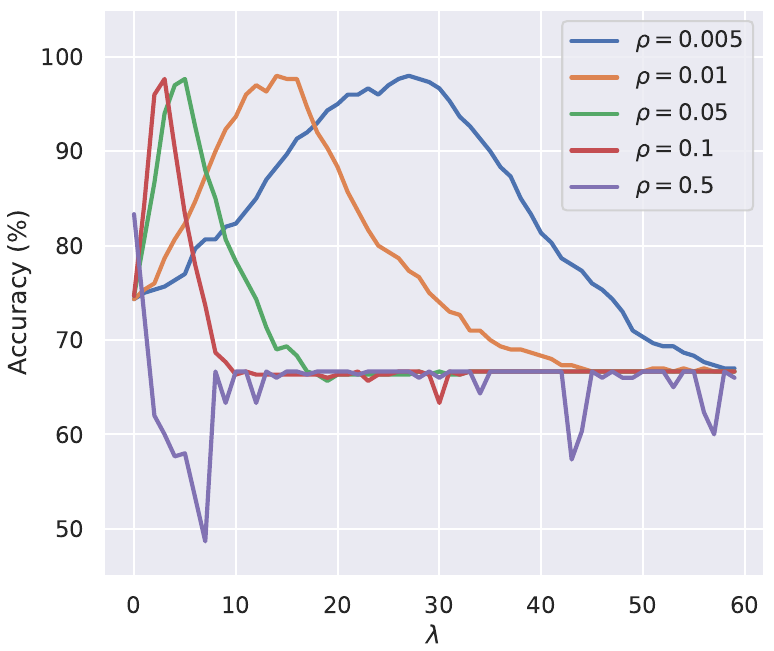}}
\vspace{-0.5em}
\caption{\textbf{Performances of classification with the varied hyperparameter $\lambda$.} The left (a) and right subplots (b) show the performance on the training and test domains, respectively.}\label{fig:lambda}
\vspace{-1.5em}
\end{figure}

\textbf{Influence of the landscape consistency on  DG performances.} 
In this experiment, we gradually increase the values of $\lambda_2$ in (\ref{eq:refine}) to strengthen the promotion of landscape consistency, while tracking the changes in DG performance.
When there's no ambiguity, we drop the subscript on $\lambda_2$ and refer to it simply as $\lambda$.
As shown in Fig.\ref{fig:lambda}, if the perturbation strength $\rho$ is enough small, the out-of-domain accuracy initially increases and then decreases as $\lambda$ increases (see Fig.\ref{fig:lambda-test}). 
This phenomenon can be explained as a trade-off between maintaining label correctness and enhancing the landscape consistency: 
Initially, with a small value of $\lambda$, the soft labels remain accurate, and increasing $\lambda$ helps improve out-of-domain generalization by encouraging landscape consistency.
However, when $\lambda$ becomes too large, it compromises the correctness of the soft labels, leading to a sharp drop in accuracy for both the training (see Fig.\ref{fig:lambda-train}) and test domains. 
Besides, as the perturbation strength $\rho$ increases, the accuracy peak on the test domain shifts to the left, which is because a larger sharpness value will be obtained with larger $\rho$, 
alleviating the need for excessively large $\lambda$ to promote consistency. Finally, when $\rho$ reaches 0.5, accuracies on the training and test domains deteriorate rapidly, similar to the behavior of the SAM~\cite{SAM}.

\subsection{Experiments on Real Dataset}
\setlength{\tabcolsep}{20pt}
\begin{table*}[htbp]
\caption{\textbf{Comparison with popular DG methods with ResNet-50 pre-trained on ImageNet.} 
The table reports the average out-of-domain accuracy on five DG datasets. Each out-of-domain performance is an average of three different runs with distinct train-validation splits.
We highlight the best results in \textbf{bold} and \underline{underline} the second best results.
Results marked by $\dagger$ and $\ddagger$ are cited from
Gulrajani and Lopez-Paz~\cite{DomainBed} and Cha \emph{et al.}~\cite{SWAD}, respectively.}
\vspace{-0.8em}
\label{Table:ResNet}
\centering
\scalebox{0.8}{\begin{tabular}{l|ccccc|c}
\toprule
    
    \textbf{Algorithms} & VLCS & PACS & OfficeHome & TerraIncognita & DomainNet & Avg. \\
    \midrule
    ERM$^\dagger$~\cite{ERM} & 77.5\scriptsize{$\pm$0.4} & 85.5\scriptsize{$\pm$0.2} & 66.5\scriptsize{$\pm$0.3} & 46.1\scriptsize{$\pm$1.8} & 40.9\scriptsize{$\pm$0.1} & 63.3  \\
    IRM$^\dagger$~\cite{IRM} & 78.5\scriptsize{$\pm$0.5} & 83.5\scriptsize{$\pm$0.8} & 64.3\scriptsize{$\pm$2.2} & 47.6\scriptsize{$\pm$0.8} & 33.9\scriptsize{$\pm$2.8} & 61.6 \\
    GroupDRO$^\dagger$~\cite{DRO} & 76.7\scriptsize{$\pm$0.6} & 84.4\scriptsize{$\pm$0.8} & 66.0\scriptsize{$\pm$0.7} & 43.2\scriptsize{$\pm$1.1} & 33.3\scriptsize{$\pm$0.2} & 60.7 \\
    Mixup$^\dagger$~\cite{Mixup} & 77.4\scriptsize{$\pm$0.6} & 84.6\scriptsize{$\pm$0.6} & 68.1\scriptsize{$\pm$0.3} & 47.9\scriptsize{$\pm$0.8} & 39.2\scriptsize{$\pm$0.1} & 63.4 \\
    MLDG$^\dagger$~\cite{MLDG} & 77.2\scriptsize{$\pm$0.4} & 84.9\scriptsize{$\pm$1.0} & 66.8\scriptsize{$\pm$0.6} & 47.7\scriptsize{$\pm$0.9} & 41.2\scriptsize{$\pm$0.1} & 63.6 \\
    CORAL$^\dagger$~\cite{CORAL} & 78.8\scriptsize{$\pm$0.6} & 86.2\scriptsize{$\pm$0.3} & 68.7\scriptsize{$\pm$0.3} & 47.6\scriptsize{$\pm$1.0} & 41.5\scriptsize{$\pm$0.1} & 64.6 \\
    MMD$^\dagger$~\cite{MMD} & 77.5\scriptsize{$\pm$0.9} & 84.6\scriptsize{$\pm$0.5} & 66.3\scriptsize{$\pm$0.1} & 42.2\scriptsize{$\pm$1.6} & 23.4\scriptsize{$\pm$9.5} & 58.8 \\
    DANN$^\dagger$~\cite{DANN} & 78.6\scriptsize{$\pm$0.4} & 83.6\scriptsize{$\pm$0.4} & 65.9\scriptsize{$\pm$0.6} & 46.7\scriptsize{$\pm$0.5} & 38.3\scriptsize{$\pm$0.1} & 62.6 \\
    CDANN$^\dagger$~\cite{cDANN} & 77.5\scriptsize{$\pm$0.1} & 82.6\scriptsize{$\pm$0.9} & 65.8\scriptsize{$\pm$1.3} & 45.8\scriptsize{$\pm$1.6} & 38.3\scriptsize{$\pm$0.3} & 62.0 \\
    MTL$^\dagger$~\cite{MTL} & 77.2\scriptsize{$\pm$0.4} & 84.6\scriptsize{$\pm$0.5} & 66.4\scriptsize{$\pm$0.5} & 45.6\scriptsize{$\pm$1.2} & 40.6\scriptsize{$\pm$0.1} & 62.9 \\ 
    SagNet$^\dagger$~\cite{Sagnet} & 77.8\scriptsize{$\pm$0.5} & \underline{86.3}\scriptsize{$\pm$0.2} & 68.1\scriptsize{$\pm$0.1} & 48.6\scriptsize{$\pm$1.0} & 40.3\scriptsize{$\pm$0.1} & 64.2 \\ 
    ARM$^\dagger$~\cite{ARM} & 77.6\scriptsize{$\pm$0.3} & 85.1\scriptsize{$\pm$0.4} & 64.8\scriptsize{$\pm$0.3} & 45.5\scriptsize{$\pm$0.3} & 35.5\scriptsize{$\pm$0.2} & 61.7 \\
    VREx$^\dagger$~\cite{VREx} & 78.3\scriptsize{$\pm$0.2} & 84.9\scriptsize{$\pm$0.6} & 66.4\scriptsize{$\pm$0.6} & 46.4\scriptsize{$\pm$0.6} & 33.6\scriptsize{$\pm$2.9} & 61.9 \\
    RSC$^\dagger$~\cite{RSC} & 77.1\scriptsize{$\pm$0.5} & 85.2\scriptsize{$\pm$0.9} & 65.5\scriptsize{$\pm$0.9} & 46.6\scriptsize{$\pm$1.0} & 38.9\scriptsize{$\pm$0.5} & 62.7 \\
    Mixstyle$^\ddagger$~\cite{Mixstyle} & 77.9\scriptsize{$\pm$0.5} & 85.2\scriptsize{$\pm$0.3} & 60.4\scriptsize{$\pm$0.3} & 44.0\scriptsize{$\pm$0.7} & 34.0\scriptsize{$\pm$0.1} & 60.3 \\
    AndMask~\cite{AndMask} & 78.1\scriptsize{$\pm$0.9} & 84.4\scriptsize{$\pm$0.9} & 65.6\scriptsize{$\pm$0.4} & 44.6\scriptsize{$\pm$0.3} & 37.2\scriptsize{$\pm$0.6} & 62.0 \\
    Fish~\cite{Fish} & 77.8\scriptsize{$\pm$0.3} & 85.5\scriptsize{$\pm$0.3} & 68.6\scriptsize{$\pm$0.4} & 45.1\scriptsize{$\pm$1.3} & 42.7\scriptsize{$\pm$0.2} & 63.9 \\
    SelfReg$^\dagger$~\cite{SelfReg} & 77.8\scriptsize{$\pm$0.9} & 85.6\scriptsize{$\pm$0.4} & 67.9\scriptsize{$\pm$0.7} & 47.0\scriptsize{$\pm$0.3} & 42.8\scriptsize{$\pm$0.0} & 64.2 \\
    mDSDI~\cite{mDSDI} & 79.0\scriptsize{$\pm$0.3} & 86.2\scriptsize{$\pm$0.2} & 69.2\scriptsize{$\pm$0.4} & 48.1\scriptsize{$\pm$1.4} & 42.8\scriptsize{$\pm$0.1} & 65.1 \\
    \midrule
    MIRO~\cite{MIRO} & 79.0\scriptsize{$\pm$0.0} & 85.4\scriptsize{$\pm$0.4} & \underline{70.5}\scriptsize{$\pm$0.4} & \underline{50.4}\scriptsize{$\pm$1.1} & \underline{44.3}\scriptsize{$\pm$0.2} & \underline{65.9} \\
    \midrule
    SAM$^\ddagger$~\cite{SAM} & \underline{79.4}\scriptsize{$\pm$0.1} & 85.8\scriptsize{$\pm$0.2} & 69.6\scriptsize{$\pm$0.1} & 43.3\scriptsize{$\pm$0.7} & \underline{44.3}\scriptsize{$\pm$0.0} & 64.5 \\
    SFT (Ours) & \textbf{79.8}\scriptsize{$\pm$0.1} & \textbf{88.3}\scriptsize{$\pm$0.3} & \textbf{70.9}\scriptsize{$\pm$0.1} & \textbf{50.7}\scriptsize{$\pm$0.4} & \textbf{46.0}\scriptsize{$\pm$0.0} & \textbf{67.1} \\
\bottomrule
\end{tabular}}
\end{table*}

\setlength{\tabcolsep}{16pt}
\begin{table*}
  \caption{\textbf{DG performances on large-scale Vision Transformers.} Out-of-domain accuracies of two backbones (ViT-B/16 and ViT-L/14) are shown below.
  The presence or absence of ``*'' indicates whether full fine-tuning or visual prompt tuning is employed, respectively.
  }
  \vspace{-0.8em}
  \label{Table:ViT}
  \centering
  \scalebox{0.8}{
  \begin{tabular}{c|l|ccccc|c}
    \toprule
    \textbf{Backbone} & \textbf{Algorithms} & VLCS & PACS & OfficeHome & TerraIncognita & DomainNet & Avg. \\
    \midrule
    \multirow{11}{*}{ViT-B/16} & ERM$^*$~\cite{ERM}  & 81.4 & 92.9 & 78.9 & 53.6 & 56.1 & 72.6 \\
    &  MIRO$^*$ ~\cite{MIRO} & 82.2 & 95.6 & 82.5 & 54.3 & 54.0 & 73.7 \\
    \cmidrule(r){2-8}
    & ERM~\cite{ERM} & 80.9 & \underline{96.6} & 84.1 & 55.5 & 59.2 & 75.3 \\
    & IRM~\cite{IRM} & 81.9 & 96.4 & 83.1 & 50.9 & 59.1 & 74.3 \\
    & DANN~\cite{DANN} & 81.7 & 95.5 & 82.7 & 52.0 & 58.6 & 74.1 \\
    & CDANN~\cite{cDANN} & 81.9 & 96.0 & 82.3 & 54.9 & 58.4 & 74.7 \\
    & CORAL~\cite{CORAL} & 82.5 & 95.4 & 83.3 & 52.0 & 59.5 & 74.5 \\
    & MMD~\cite{MMD} & 81.9 & 95.1 & 83.7 & 56.9 & \underline{59.9} & 75.5 \\
    & IIB~\cite{IIB} & 82.5 & 96.0 & 83.9 & \underline{58.0} & 58.6 & 75.8 \\
    \cmidrule(r){2-8}
    & SAM~\cite{SAM} & \underline{83.5} & 96.1 & \underline{85.7} & 56.6 & 59.8 & \underline{76.3} \\
    & SFT (Ours) & \textbf{84.1} & \textbf{96.8} & \textbf{86.5} & \textbf{61.2} & \textbf{60.5} & \textbf{77.8} \\
    \midrule
    \multirow{3}{*}{ViT-L/14} & ERM~\cite{ERM} & 82.9 & \textbf{98.8} & 90.2 & 61.3 & \underline{65.4} & 79.7 \\
    & SAM~\cite{SAM} & \underline{84.2} & \underline{98.7} & \underline{90.8} & \underline{62.8} & 65.2 & \underline{80.3} \\
    & SFT (Ours) & \textbf{84.4} & 98.6 & \textbf{91.3} & \textbf{65.2} & \textbf{66.5} & \textbf{81.2} \\
    \bottomrule
  \end{tabular}}
  \vspace{-0.7em}
\end{table*}
\subsubsection{Experimental setting}
\textbf{Dataset and protocol.} We evaluate our method on five widely-used and challenging datasets for domain generalization, including PACS~\cite{PACS} (4 domains, 7 classes, and 9, 991 examples), VLCS~\cite{VLCS} (4 domains, 5 classes, and 10, 729 examples), OfficeHome~\cite{OH} (4 domains, 65 classes, and 15, 588 images), TerraIncognita~\cite{TI} (4 domains, 10 classes, and 24, 788 examples) and DomainNet~\cite{DN} (6 domains, 345 classes, and 586, 575 examples).
Our experimental environment is built based on the well-known DomainBed
benchmark~\cite{DomainBed}. Following the training and evaluation protocol of DomainBed, we select one domain for testing and the remaining domains for training every time, and 20\% samples of training domains are held out for validation and model selection. Model selection is carried out based on the training-domain validation set. 

\textbf{Implementation details.} To illustrate the broad applicability of the proposed method, we conduct experiments based on small models (ImageNet pre-trained ResNet-50) and large-scale pre-trained models (ViT-B/16 and ViT-L/14 for CLIP
~\cite{CLIP}). 
Unless otherwise specified, the landscape refiner shares the same architecture with the model $\boldsymbol{f}_{\boldsymbol{\theta}}$ in this paper.
Through random hyperparameter search, we determine the batch size, learning rates, and other hyperparameters. The detailed search spaces for each are provided in the \textbf{supplementary file}. The Adam~\cite{Adam} optimizer is utilized in all of our experiments.

\subsubsection{Main Results}
\textbf{Experiments with the ResNet-50 backbone.}
We begin by comparing SFT with conventional DG methods. As shown in Table \ref{Table:ResNet}, SFT outperforms ERM~\cite{ERM} across all five datasets, showing an average improvement of 3.8\%.
Furthermore, SFT consistently outperforms other conventional DG methods, including invariant or causal representation learning methods (IRM~\cite{IRM}, VREx~\cite{VREx}, CORAL~\cite{CORAL}, MMD~\cite{MMD}, DANN~\cite{DANN}, CDANN~\cite{cDANN}), disentangled representation learning methods (SagNet~\cite{Sagnet} and mDSDI~\cite{mDSDI}), distributionally robust optimization methods (GroupDRO~\cite{DRO}), meta-learning methods (MLDG~\cite{MLDG}, ARM~\cite{ARM}, and Fish~\cite{Fish}), data or feature augmentation methods (Mixup~\cite{Mixup} and Mixstyle~\cite{Mixstyle}), and other methods (RSC~\cite{RSC}, AndMask~\cite{AndMask}, MTL~\cite{MTL}, and SelfReg~\cite{SelfReg}). As a strong baseline, SAM achieves an average improvement of +1.2\% (63.3\% $\rightarrow$ 64.5\%) over ERM and outperforms most other baselines, except for mDSDI~\cite{mDSDI} and MIRO~\cite{MIRO}. By refining loss landscapes to improve consistency, SFT provides an additional average gain of +2.6\% (64.5\% $\rightarrow$ 67.1\%) over SAM. Remarkably, our SFT even outperforms the robust  MIRO~\cite{MIRO} baseline by 1.2\%.

\textbf{Experiments with Large-scale Vision Transformers (ViTs).}
To demonstrate the versatility of SFT across different architectures, we also conduct experiments using large-scale ViTs. 
Given the efficiency of Visual Prompt Tuning (VPT)~\cite{VPT}—which requires tuning only 1\% of the parameters—and its strong performance in domain generalization, as shown in Table \ref{Table:ViT} and previous works~\cite{CSVPT,DoPrompt}, we adopt VPT in our ViT experiments.
From Table \ref{Table:ViT}, we can see that SFT achieves superior performance over conventional DG methods and the vanilla SAM (+1.5\% average improvement) when using ViT-B/16. When using a larger pre-trained ViT-L/14, the out-of-domain accuracies further improve, achieving 79.7\% for ERM, 80.3\% for SAM, and 81.2\% for SFT. 
The results highlight that SFT can be easily integrated with popular tuning methods like VPT, and there is potential for further enhancement with advanced tuning techniques in future work.

\textbf{Comparison with popular sharpness-aware methods.} 
We also perform a more fine-grained comparison of SFT with popular SAM variants designed to seek better flat minima. As shown in Table \ref{Table:SAM}, SFT generally outperforms these methods across all five datasets, with average improvements of  2.7\%, 2.0\%, 1.8\% and 1.0\% over GAM~\cite{GAM}, GSAM~\cite{GSAM}, FAD~\cite{FAD} and SAGM~\cite{SAGM}, respectively. These experimental results further highlight the importance of the consistency of flat minima.
Since SFT currently only applies the basic SAM algorithm, there is potential for further improvements by incorporating more advanced SAM variants, such as SAGM~\cite{SAGM}, which remains a focus for our future research.

\setlength{\tabcolsep}{5pt}
\begin{table}
  \caption{\textbf{Comparisons of SFT and popular sharpness-aware methods.} This table presents average DG performances on five datasets, and the mean and standard deviation for each dataset.
  }
  \label{Table:SAM}
  \centering
  \scalebox{0.65}{\begin{tabular}{l|ccccc|c}
   \toprule
   \textbf{Algorithms} & VLCS & PACS & OfficeHome & TerraIncognita & DomainNet & Avg. \\
   \midrule
    SAM~\cite{SAM} & 79.4\scriptsize{$\pm$0.1} & 85.8\scriptsize{$\pm$0.2} & 69.6\scriptsize{$\pm$0.1} & 43.3\scriptsize{$\pm$0.7} & 44.3\scriptsize{$\pm$0.0} & 64.5 \\
    GAM~\cite{GAM} & 78.5\scriptsize{$\pm$0.4} & 86.1\scriptsize{$\pm$0.6} & 68.2\scriptsize{$\pm$1.0} & 45.2\scriptsize{$\pm$0.6} & 43.8\scriptsize{$\pm$0.1} & 64.4 \\
    GSAM~\cite{GSAM} & 79.1\scriptsize{$\pm$0.2} & 85.9\scriptsize{$\pm$0.1} & 69.3\scriptsize{$\pm$0.0} & 47.0\scriptsize{$\pm$0.8} & 44.6\scriptsize{$\pm$0.2} & 65.1 \\
    FAD~\cite{FAD} & 78.9\scriptsize{$\pm$0.8} & \underline{88.2}\scriptsize{$\pm$0.5} & 69.2\scriptsize{$\pm$0.5} & 45.7\scriptsize{$\pm$1.0} & 44.4\scriptsize{$\pm$0.1} & 65.3 \\
    SAGM~\cite{SAGM} & \textbf{80.0}\scriptsize{$\pm$0.3} & 86.6\scriptsize{$\pm$0.2} & \underline{70.1}\scriptsize{$\pm$0.2} & \underline{48.8}\scriptsize{$\pm$0.9} & \underline{45.0}\scriptsize{$\pm$0.2} & \underline{66.1} \\
    SFT (Ours) & \underline{79.8}\scriptsize{$\pm$0.1} & \textbf{88.3}\scriptsize{$\pm$0.3} & \textbf{70.9}\scriptsize{$\pm$0.1} & \textbf{50.7}\scriptsize{$\pm$0.4} & \textbf{46.0}\scriptsize{$\pm$0.0} & \textbf{67.1} \\
    \bottomrule
  \end{tabular}}
\end{table}

\subsection{Sharpness Analysis}
\begin{figure}[t]
\centering
\includegraphics[scale=0.24]{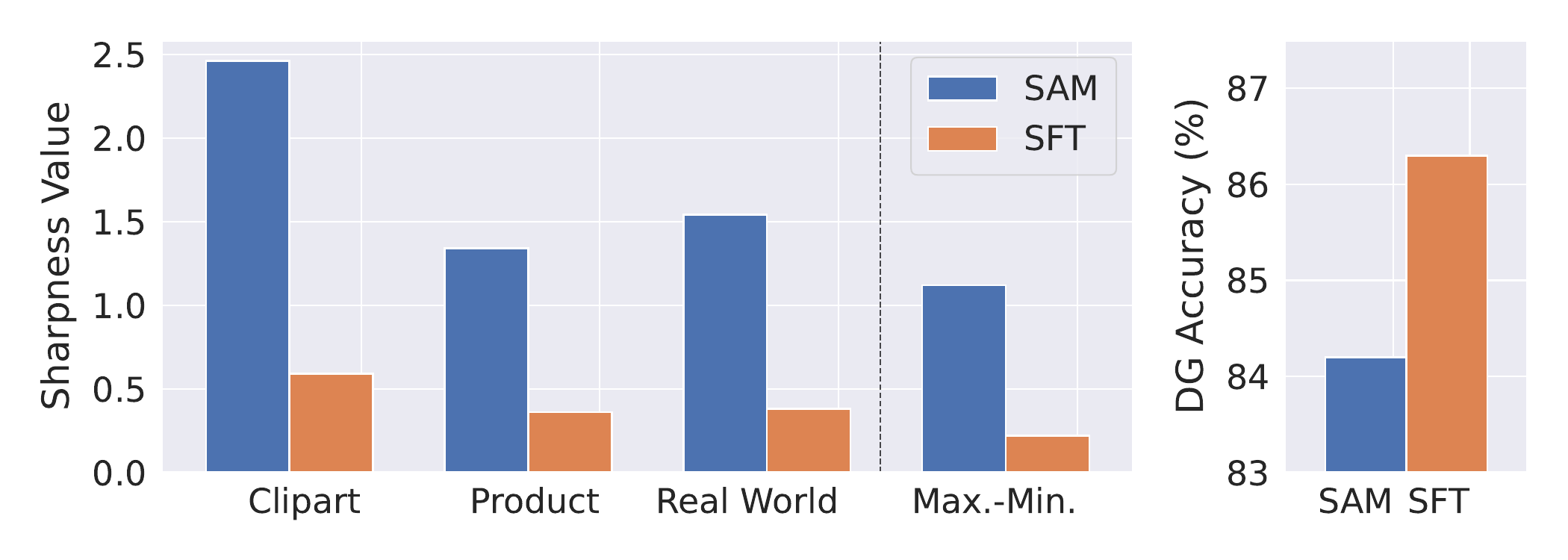}
\vspace{-0.8cm}
\caption{\textbf{Comparison of model sharpness and DG accuracy across different training strategies on the OfficeHome dataset.}}\label{fig:sharp}
\end{figure}

We also perform a quantitative analysis of the landscape consistency. As mentioned earlier, SFT quantifies landscape consistency through domain-specific sharpness differences. In Figure \ref{fig:sharp}, we find that SFT reduces domain-specific sharpness values and their differences compared to SAM, while simultaneously attaining higher classification accuracy. These empirical results confirm that SFT effectively improves the loss landscape consistency across domains, while also demonstrating the positive impact of landscape consistency on the DG performance.

\subsection{Ablation Study}
Ablation studies on the key components of the SFT framework are summarized in Table \ref{Table:Ablation}. The upper part of the table shows experimental results for ERM and SAM with one-hot and smoothed labels, while the lower part presents results using soft labels generated by the landscape refiner. The results in the upper part confirm the effectiveness of SAM (where $\rho\neq 0$) and label smoothing. 
When comparing the six experiments in the lower part, we observe that incorporating all components of the SFT framework to enhance  landscape consistency across domains yields the best performance, achieving an average accuracy of 86.5\%. Furthermore, removing any component from the SFT framework results in a performance drop, which underscores the importance of each element within the framework.
Finally, the comparison between the upper and lower sections demonstrates that the consistency-guided soft labels in our SFT framework offer a clear advantage over traditional one-hot or smoothed labels, significantly improving model generalization.


\setlength{\tabcolsep}{5pt}
\begin{table}[ht]
	\caption{\textbf{Ablative experiments on the OfficeHome dataset.} The ``Label'' column specifies the type of labels used, such as one-hot labels, smoothed labels or labels generated by the landscape refiner. The ``PCE'' indicates whether the PCE loss is utilized in the Exp.5-10. If not,  we use the standard cross entropy with label smoothing instead.
  }
	\label{Table:Ablation}
        \vspace{-0.3cm}
	\centering
	\scalebox{0.6}{\begin{tabular}{c|ccccc|cccc|c}
			\toprule
			\textbf{Exp.} & $\rho$ & Label & PCE & $\lambda_{1}$ & $\lambda_{2}$ & Art & Clipart & Product & Real World & Avg. \\
			\midrule 
			1 & 0.0 & One-hot & $-$ & $-$ & $-$ & 83.2 & 73.8 & 90.1 & 89.1 & 84.0 \\
			2 & 0.0 & smoothed & $-$ & $-$ & $-$ & 83.6 & 74.1 & 90.0 & 89.6 & 84.3 \\
			3 & 0.3 & One-hot & $-$ & $-$ & $-$ & 83.8 & 76.6 & 90.1 & 90.3 & 85.2 \\
                4 & 0.3 & Smoothed & $-$ & $-$ & $-$ & 84.2 & 76.4 & 90.6 & 90.5 &  85.4\\
			\midrule
               5 & 0.0 & Generated & w/o & 0.0 & 0.0 & 83.9 & 74.9 & 89.9 & 90.0 & 84.7 \\
			6 & 0.3 & Generated & w/o & 0.0 & 0.0 & 84.0             & 75.9                & 90.1     & 90.1               & 85.0 \\
			7 & 0.3 & Generated & w/o & 0.5 & 0.0 & 84.8             & 76.8              & 90.3              & 90.6   & 85.6 \\
			8 & 0.3 & Generated & w/o & 0.0 & 0.7 & 84.9             & \underline{77.3}    & 90.6             & 90.5               & 85.8 \\
			9 & 0.3 & Generated & w/o & 0.5 & 0.7 & \underline{85.2} & 77.2               & \underline{90.8}  & \underline{90.7}      & \underline{86.0} \\
			10 & 0.3 & Generated & w/  & 0.5 & 0.7 & \textbf{86.3}    & \textbf{77.7}       & \textbf{91.2}              & \textbf{90.9}   & \textbf{86.5} \\
			\bottomrule   
	\end{tabular}}
        \vspace{-0.5cm}
\end{table}

\section{Conclusion} 
This paper has proposed an iterative two-phase Self-Feedback Training (SFT) framework, aiming at addressing the issue of landscape inconsistency brought by domain shifts.
It alternatively generates a feedback signal by measuring the inconsistency of loss landscapes in different domains during the feedback phase, and then refines these loss landscapes for greater consistency using this feedback signal in the refinement phase.
Benefiting from the consistency of the sought flat minima, SFT demonstrates superior performances over sharpness-aware methods and prevalent DG methods in diverse experimental settings. 
We hope that this study will inspire more research on advanced DG methods from the view of loss landscapes. Further theoretical analysis and technical improvement will be our future work.


\clearpage
\setcounter{page}{1}
\maketitlesupplementary
Let us start with a brief overview of this supplementary material. In Section \ref{section:1}, we perform a theoretical analysis of the loss landscape consistency under the PAC-Bayesian framework. The analysis shows that the sharpness of the test loss can be constrained by that of the training domains, which also implies the generalization of consistent flat minima sought by the self-feedback training (SFT). 
In Section \ref{section:2}, we discuss the proposed projection cross-entropy loss and the algorithm used to solve the associated KL divergence minimization problem. We provide a comprehensive derivation of the algorithm, along with a time comparison against an implementation using the common convex optimization library. Subsequently, we present the full results obtained using Resnet-50, ViT-B/16, and ViT-L/14 in the following two sections (Section \ref{section:3} and \ref{section:4}). Additionally, we offer necessary explanations of the code (in Section \ref{section:5}) and hyperparameters (in Section \ref{section:6}) required for reproducibility. Finally, in Section \ref{section:7}, we assure that this research is unlikely to have any significant negative social impact.

\section{Theoretical Analysis}\label{section:1}
In this section, we will perform a theoretical analysis of the loss landscape consistency under the PAC-Bayesian framework. For clarity and ease of understanding, we first provide a detailed explanation of the relevant notations and concepts that will be used throughout the analysis.

\subsection{Notations}\label{sec:notation}
Let $\mathcal{X}$ and $\mathcal{Y}$ denote the input sample space and the category space, respectively. 
Consider a dataset drawn from $p$ training distributions $\{\mathcal{D}_d\}_{d=1}^p$,
each defined over the joint space $\mathcal{X} \times \mathcal{Y}$.
Let $D_d =\{(\boldsymbol{x}_i^{(d)},y_i^{(d)})\}_{i=1}^{n_d}$ denote the dataset sampled from the $d$-th distribution $\mathcal{D}_d$, which is referred to as the $d$-th domain. 
$(\boldsymbol{x}_i^{(d)},y_i^{(d)}) \in \mathcal{X} \times \mathcal{Y}$ denotes the $i$-th sample from domain $\mathcal{D}_d$, with $n_d$ indicating the number of samples in the $d$-th domain. 
For convenience, we also use $z_i$ to denote $(\boldsymbol{x}_i,y_i)$.
Let $\Omega$ and $\Omega'$ denote the dataset space and distribution space, respectively. 
In our analysis, domain shifts are modeled by a mapping function $\omega:\Omega \rightarrow \Omega'$, which maps one dataset to another distribution with distinct statistical properties. We assume that the domain shifts $\omega$ follow a specific distribution $\mathcal{W}$.

Our SFT framework mainly involves a model and a landscape refiner. 
Let $\mathcal{H}_{m}$ and $\mathcal{H}_{r}$ denote the hypothesis spaces of the model and the refiner, respectively. 
To analyze the SFT within the PAC-Bayesian framework, we need to provide a description using Bayesian terms.
Let $\mathcal{M}_{m}$, $\mathcal{M}_{r}$ denote the sets of distributions over $\mathcal{H}_{m}$ and $\mathcal{H}_{r}$.
In order to obtain a model from its prior distribution, we first sample a prior distribution $P$ for the refiner from the hyper-prior distribution $\mathcal{P}$, which is independent of training samples. Then, we use a mapping function $\psi: \mathcal{M}_{r}\rightarrow \mathcal{M}_{m}$ to obtain the model's prior distribution $\psi(P) \in \mathcal{M}_{m}$ and sample a model $f$ from $\psi(P)$.
In order to obtain a model from its posterior distribution, we sample a posterior distribution $P$ for the refiner from a hyper-posterior distribution $\mathcal{Q}$, which may depend on training samples. Then, we apply the training algorithm to the dataset $D_d \in \Omega$  to obtain the model's posterior distribution $\mathcal{A}(D_{d}, P)$, where $\mathcal{A}: \Omega \times \mathcal{M}_{r} \rightarrow \mathcal{M}_{m}$ represents the function that maps one dataset along with the refiner's posterior distribution to the model's posterior distribution.  Finally, we can sample a model $f$ from this posterior distribution.

\subsection{Main Theorem}

In this subsection, we first introduce a lemma that will be used multiple times in our analysis, and then formally present the main result, which is stated as Theorem \ref{th:main}.

\begin{lemma}[McAllester’s bound~\cite{McAllester}]\label{lemma:PACBayes}
Let $\mathcal{X}$ be a sample space and $\mathcal{H}$ a hypothesis space of functions over $\mathcal{X}$. Given $\pi$ be some prior distribution over hypothesis space $\mathcal{H}$, for bounded loss $\ell: \mathcal{H}\times\mathcal{X}\rightarrow [0,1]$ and any $\delta \in (0, 1]$, the following bound holds
uniformly for all posterior distributions $\rho$ with probability at least $1-\delta$:
\begin{align}
\underset{\theta\sim\rho}{\mathbb{E}}\ell(\theta,\mathcal{D}) \leq \underset{\theta\sim\rho}{\mathbb{E}}\ell(\theta,S_n) + \sqrt{\frac{\mathrm{KL}(\rho||\pi) + \log{(n/\delta)}}{2(n-1)}},
\end{align}
where $$\ell(\theta,\mathcal{D})=\mathbb{E}_{z\sim D}\ell(\theta,z)$$ and $$\ell(\theta,S_n) = \frac{1}{n}\sum_{i=1}^{n}\ell(\theta,z_i)$$ denotes the population loss and training loss, respectively. $S_n$ represents a dataset with $n$ training samples drawn independently and identically from distribution $\mathcal{D}$.
\end{lemma}

\begin{theorem}\label{th:main}
Consider the domain generalization problem with $p$ training domains. For each training domain $ \mathcal{D}_d $, we are given $ p-1 $ training-domain pairs $ (\mathcal{D}_d, \mathcal{D}_{d'}) $, where $d'\neq d$, with  each pair consisting of a training dataset $D_{d}$ of size $n_d$ and a hold-out dataset $D_{d'}$ of size $n_{d'}$.
Let the dataset space be $ \Omega $ and the distribution space be $ \Omega' $.
Assume that the domain shifts, denoted by $ \omega: \Omega \rightarrow \Omega' $, follow a distribution $ \mathcal{W} $.
The difference in loss sharpness between domains $ \mathcal{D}_d $ and $ \mathcal{D}_{d'} $ is defined as:
\begin{align}
    \Delta \ell (f, \mathcal{D}_d, \mathcal{D}_{d'}):= |\underset{z\sim \mathcal{D}_d}{\mathbb{E}} \ell(f,z) - \underset{z\sim \mathcal{D}_{d'}}{\mathbb{E}} \ell(f,z)|,
\end{align}
where $ \ell(f, z) $ represent the loss sharpness of the model $ f $ evaluated on data point $ z $.
Let $ \mathcal{P} $ denote a predefined hyper-prior distribution over the set of all possible prior distributions for the landscape refiner. 
Then, for all hyper-posterior distributions $ \mathcal{Q} $ that ensure a sufficiently small sharpness difference between training domains, i.e.,
\begin{align}
\forall d'\neq d: \underset{P \sim \mathcal{Q}}{\mathbb{E}}\ \underset{f \sim \mathcal{A}(D_d,P)}{\mathbb{E}} 
\Delta \ell(f,\mathcal{D}_{d}, \mathcal{D}_{d'})
\leq \epsilon,
\end{align}
and for any $ \delta \in (0, 1] $,
the following inequality holds with probability at least $ 1 - \delta $:
\begin{align}
&\quad \ell(\mathcal{Q}, \omega) \leq \epsilon +  \hat{\ell}(\mathcal{Q}, D_d)
+\frac{1}{p-1}\times\nonumber\\
& \sum_{d'\neq d}\sqrt{\frac{\mathrm{KL}(\mathcal{Q} \| \mathcal{P})+\underset{P \sim \mathcal{Q}}{\mathbb{E}} \mathrm{KL}\left(\mathcal{A} \| \psi(P)\right)+\log \frac{2(p-1) n_d}{\delta}}{2(n_d-1)}} \nonumber \\
&+\sqrt{\frac{\mathrm{KL}(\mathcal{Q} \| \mathcal{P})+\log \frac{2(p-1)}{\delta}}{2(p-2)}}.
\end{align}
Here, $ \psi(P)$ is the prior model distribution, and $ \mathcal{A}$ is a shorthand of the posterior model distribution $ \mathcal{A}(D_d, P) $. $ \mathrm{KL}(\cdot \| \cdot) $ denotes the Kullback-Leibler divergence.
The term $ \ell(\mathcal{Q}, \omega) $ is defined as:
\begin{align}  
\ell(\mathcal{Q}, \omega) :=
\underset{P \sim \mathcal{Q}}{\mathbb{E}}\ 
\underset{\omega \sim \mathcal{W}}{\mathbb{E}}\ 
\underset{f \sim \mathcal{A}(D_d, P)}{\mathbb{E}}\ 
\underset{z \sim \omega(D_{d})}{\mathbb{E}} \ell(f, z),
\end{align}
which represents the expected sharpness of the model evaluated on the test distribution $ \omega(D_d) $.
The term $ \hat{\ell}(\mathcal{Q}, D_d) $ is defined as:
\begin{align}  
\hat{\ell}(\mathcal{Q}, D_{d}) :=
\underset{P \sim \mathcal{Q}}{\mathbb{E}}\ 
\underset{f \sim \mathcal{A}(D_d,P)}{\mathbb{E}}
\frac{1}{n_d} \sum_{i=1}^{n_d}  \ell\left(f, z_{i}\right),
\end{align}
which represents the empirical sharpness of the model over the training domain $ D_d $.
\end{theorem}

\begin{proof}
Firstly, we bound the loss sharpness in each of the domain pairs.
Based on the above definitions in the subsection \ref{sec:notation}, we can decompose the KL divergence term of Lemma \ref{lemma:PACBayes} in the following way:
\begin{align}
&\quad \mathrm{KL}(\rho \| \pi) \nonumber\\
&=\underset{f \sim \rho}{\mathbb{E}} \log \frac{\rho(f)}{\pi(f)} =\underset{P \sim \mathcal{Q}}{\mathbb{E}}\ \underset{f \sim \mathcal{A}(D_d, P)}{\mathbb{E}} \log \frac{\mathcal{Q}(P) \mathcal{A}\left(D_d, P\right)(f)}{\mathcal{P}(P) \psi(P)(f)} \nonumber\\
& =\underset{P \sim \mathcal{Q}}{\mathbb{E}} \log \frac{\mathcal{Q}(P)}{\mathcal{P}(P)} +
\underset{P \sim \mathcal{Q}} {\mathbb{E}}\ 
\underset{f \sim \mathcal{A}(D_d, P)}{\mathbb{E}}
\log \frac{\mathcal{A}\left(D_d, P\right)(f)}{\psi(P)(f)} \nonumber\\
& =\mathrm{KL}(\mathcal{Q} \| \mathcal{P})+\underset{P \sim \mathcal{Q}}{\mathbb{E}} \mathrm{KL}\left(\mathcal{A}\left(D_d, P\right) \| \psi(P)\right).
\end{align}
This decomposition separates the KL divergence into two components: $\mathrm{KL}(\mathcal{Q} \| \mathcal{P})$ and $\underset{P \sim \mathcal{Q}}{\mathbb{E}} \mathrm{KL}\left(\mathcal{A}\left(D_d, P\right) \| \psi(P)\right)$. Now, based on this, we can establish a probabilistic upper bound on the loss sharpness.
Thus, with probability at least $1-\delta_{d'}$, we have:
\begin{align}
& \quad\underset{P \sim \mathcal{Q}}{\mathbb{E}}\  
\underset{f \sim \mathcal{A}\left(D_d, P\right)}{\mathbb{E}}\ 
\underset{z \sim \mathcal{D}_{d'}}{\mathbb{E}} 
\ell(f, z) \nonumber\\
&\leq
\underset{P \sim \mathcal{Q}}{\mathbb{E}}\ 
\underset{f \sim \mathcal{A}\left(D_d, P\right)}{\mathbb{E}}
\frac{1}{n_d} \sum_{i=1}^{n_d}  \ell\left(h, z_{i}\right) \nonumber\\ 
&\quad+\underset{P \sim \mathcal{Q}}{\mathbb{E}}\ 
\underset{f \sim \mathcal{A}\left(D_d, P\right)}{\mathbb{E}} \Delta \ell(h,\mathcal{D}_d, \mathcal{D}_{d'})
\nonumber\\
&\quad+\sqrt{\frac{\mathrm{KL}(\mathcal{Q} \| \mathcal{P})+\underset{P \sim \mathcal{Q}}{\mathbb{E}} \mathrm{KL}\left(\mathcal{A}\| \psi(P)\right)+\log \frac{n_d}{\delta_{d'}}}{2\left(n_d-1\right)}}.
\end{align}
This bound captures the generalization ability between training domains, i.e., from domain $\mathcal{D}_d$ to domain $\mathcal{D}_{d'}$.

Next, assuming that the domain shifts $ \omega: \Omega \rightarrow \Omega' $ are governed by a distribution $ \mathcal{W} $, we can
apply Lemma \ref{lemma:PACBayes}
again to bound the loss sharpness on the test domain $\omega(D_d)$. This step is critical because it extends the generalization to the test domains. Specifically, with probability at least $1-\delta'$, we obtain the following bound:
\begin{align}
&\quad\underset{P \sim \mathcal{Q}}{\mathbb{E}}\ 
\underset{\omega \sim \mathcal{W}}{\mathbb{E}}\  
\underset{f \sim \mathcal{A}(D_d, P)}{\mathbb{E}}\ 
\underset{z \sim \mathcal \omega(D_d)}{\mathbb{E}} \ell(f, z) \nonumber\\
&\leq
\underset{P \sim \mathcal{Q}}{\mathbb{E}}
\frac{1}{p-1} \sum_{d'\neq d} 
\underset{f \sim \mathcal{A}(D_d, P)}{\mathbb{E}}\ 
\underset{z \sim \mathcal{D}_{d'}}{\mathbb{E}} \ell(f, z) \nonumber\\
&+\sqrt{\frac{\mathrm{KL}(\mathcal{Q} \| \mathcal{P})+\log \frac{p-1}{\delta'}}{2(p-2)}}.
\end{align}
Here, the first term represents an average over all domain pairs, while the second term corresponds to the KL divergence between hyper-prior  and hyper-posterior distributions. The confidence parameter $\delta'$ controls the probability of the bound holding.

Finally, we set $\delta'=\delta/2$ and $\delta_{d'}=\delta/[2(p-1)]$ and apply  the union bound to combine the above results. This leads to the final bound, which holds with probability at least $1-\delta$,
\begin{align}
&\quad\underset{P \sim \mathcal{Q}}{\mathbb{E}}\ 
\underset{\omega \sim \mathcal{W}}{\mathbb{E}}\ 
\underset{f \sim \mathcal{A}(D_d, P)}{\mathbb{E}}\ 
\underset{z \sim \omega(D_d)}{\mathbb{E}} \ell(h, z) \nonumber\\
&\leq
\frac{1}{p-1} \sum_{d'\neq d} 
\left[
\underset{P \sim \mathcal{Q}}{\mathbb{E}}\  
\underset{f \sim \mathcal{A}\left(D_d, P\right)}{\mathbb{E}}
\frac{1}{n_d} \sum_{i=1}^{n_d}  \ell\left(f, z_{i}\right)\right. \nonumber\\ 
&\left.+\underset{P \sim \mathcal{Q}}{\mathbb{E}}\  
\underset{f \sim \mathcal{A}\left(D_d, P\right)}{\mathbb{E}} \Delta \ell(f,\mathcal{D}_d, \mathcal{D}_{d'})\right.+
\nonumber\\
&\left.\sqrt{\frac{\mathrm{KL}(\mathcal{Q} \| \mathcal{P})+\underset{P \sim \mathcal{Q}}{\mathbb{E}} \mathrm{KL}\left(\mathcal{A} \| \psi(P)\right)+\log \frac{2(p-1) n_d}{\delta}}{2(n_d-1)}}
\right] \nonumber\\
&+\sqrt{\frac{\mathrm{KL}(\mathcal{Q} \| \mathcal{P})+\log \frac{2(p-1)}{\delta}}{2(p-2)}}.
\end{align}
The inequality above provides a comprehensive bound of the loss sharpness on the test domains, completing the proof of Theorem \ref{th:main}.
\end{proof}

In conclusion, this theorem shows that if domain shifts are governed by a specific distribution and the domain shifts in the training domains are independently and identically sampled from this distribution, then the sharpness of the test loss is bounded by the sharpness observed in the training domains with high probability. In other words, the consistency of the flat minima achieved in the training domains can be transferred to unseen test domains. As a result, the model can exhibit strong generalization performance when applied to test domains.

\section{Projection Cross Entropy}\label{section:2}

As mentioned in the main text, the projection cross entropy (PCE) can be  used as a loss term to maintain the label correctness during the refinement phase. An efficient algorithm (Alogrithm \ref{alg:PCE}) has been presented to address the associated KL divergence minimization problem there.   

In this section, we first provide the derivation of this algorithm, followed by a comparison of its time efficiency with that of widely-used convex optimization libraries. 
Finally, we further discuss the connections between the PCE loss and other related loss functions.

\subsection{KL Divergence Minimization}
As stated in the main text, the optimization is formulated as:
\begin{equation}\label{eq:KL-9}
\min_{\mathbf{y}} \, \mathrm{KL}(\mathbf{y} \, || \, \Tilde{\mathbf{y}}) \quad \text{subject to} \quad \mathbf{y} \in C_1,
\end{equation}
where $\Tilde{\textbf{y}}$ represents the soft label output by the landscape refiner, and the label space $C_1$ can be expressed as:
\begin{equation}
C_1 = \{(q_1,\ldots,q_N) \mid \forall k \neq 1: q_{1} \geq \alpha q_k, \sum_{k=1}^{N} q_k = 1\}.
\end{equation}
Here, $\alpha \geq 1$ is a hyperparameter that controls the minimum ratio between $q_1$ and $q_k$, while $N$ denotes the number of categories for classification.
For ease of understanding, we restate the problem more explicitly as follows:
\begin{align}
	&\min\limits_{q_i} \sum_{i=1}^{N}  q_i \log \frac{q_i}{p_i} \nonumber \\
	\text{s.\ t.}\quad &\sum_{i=1}^N q_i=1,\quad q_1\geq \alpha q_k\ ( k\neq 1), \label{op}
\end{align}
where we use $(p_1,\ldots,p_N)$  to represent $\Tilde{\textbf{y}}$ for clarity.

In the following, we will offer a detailed derivation of Algorithm \ref{alg:PCE}, which is capable of finding the exact optimal solution to this optimization problem. The general idea of the derivation is as follows: first, by examining the Lagrangian function and the Karush-Kuhn-Tucker (KKT) conditions of the problem, we obtain the general form of the optimal solution. Then, we determine which constraints among the inequality constraints hold as a strict equality (i.e., active), and finally, we find the optimal solution based on these active constraints.

\subsubsection{Lagrangian Function and KKT Conditions}

To solve the given optimization problem, we first construct the Lagrangian function by incorporating the objective function and the constraints using Lagrange multipliers:
\begin{align}
L(q_i,\mu_k,\lambda) &= \sum_{i=1}^{N} q_i \log \frac{q_i}{p_i}
+ \sum_{k=2}^{N} \mu_k (\alpha q_k -q_1) \nonumber \\
&+ \lambda (\sum_{i=1}^N q_i-1),
\end{align}
\\
where $\mu_k \geq 0$ represents the Lagrange multiplier for the inequality constraint $q_1 \geq \alpha q_k$, and $\lambda$ denotes the Lagrange multiplier for the equality constraint $\sum_{i=1}^N q_i = 1$.
Then, the Karush-Kuhn-Tucker (KKT) conditions are the necessary conditions for optimality in constrained optimization problems. We apply these conditions to the Lagrangian function and derive the following set of equations:
\begin{enumerate}
    \item Stationarity. \
    The stationarity conditions require that the partial derivatives of the Lagrangian with respect to each of the variables be zero, which corresponds to the optimality condition. For $q_1$, we have the following equation:
    \begin{equation} 
        \frac{\partial L}{\partial q_1} = 1+\log{q_1}-\log{p_1}- \sum_{k=2}^{N} \mu_k + \lambda = 0.
    \end{equation}
    Similarly, for $q_j$, where $j = 2, \ldots, n$, we get:
    \begin{equation}
        \frac{\partial L}{\partial q_j} = 1+\log{q_j}-\log{p_j}+ \mu_j \alpha + \lambda = 0.
    \end{equation}
These two equations can be solved to express $q_1$ and $q_j$ in terms of the Lagrange multipliers $\mu_k$ and $\lambda$. Thus, we have the following solutions:
\begin{equation}\label{KL_x1}
    q_1 = p_1 \exp{\left(\sum_{k=2}^N \mu_k - 1 - \lambda \right)}
\end{equation}
and
\begin{equation}\label{KL_xj}
    q_j = p_j \exp{\left(-\mu_j \alpha - 1 - \lambda \right)}.
\end{equation}

\item Primal feasibility. The primal feasibility condition ensures that the original constraints are satisfied. Therefore, we have:
\begin{equation}
    \alpha q_j - q_1 \leq 0.
\end{equation}

\item Dual feasibility. The dual feasibility condition imposes non-negativity on the Lagrange multipliers associated with the inequality constraints:
\begin{equation} \label{KL_dual}
    \mu_j \geq 0.
\end{equation}

\item Complementary slackness. 
Finally, the complementary slackness condition relates the primal and dual variables. In this case, the complementary slackness condition for the inequality constraint is:
\begin{equation}
    \mu_j (\alpha q_j - q_1) = 0.
\end{equation}
This condition implies that either $\mu_j = 0$ or $\alpha q_j = q_1$ (or both). 
It ensures that if the constraint is inactive (i.e., it holds as a strict inequality), the corresponding multiplier is zero, and if the multiplier is positive,
the corresponding constraint is active (i.e., it holds as a strict equality).
\end{enumerate}

\vspace{0.2cm}
In the following, we can obtain the general form of the optimal solution by examining the two cases of $\mu > 0$ and $\mu = 0$ as per the KKT conditions.
\begin{itemize}
    
    \item \textbf{Case 1:} $\mu_j > 0$.\\

   If $\mu_j > 0$, from the stationarity condition and complementary slackness, we can derive the following equation:
    \begin{align} \label{KL=}
         &\alpha q_j = \alpha p_j \exp{(-\mu_j \alpha -1-\lambda)} \nonumber\\
        =& q_1=p_1\exp{\left(\sum_{k=2}^N \mu_k-1-\lambda\right)}.
    \end{align}
    Through a simple manipulation of the equation, we can obtain:
    \begin{equation}
        \frac{p_1}{\alpha p_j} = \frac{\exp(-\mu_j\alpha)}{\exp(\sum_{k=2}^N \mu_k)}.
    \end{equation}
    Since $\mu_j > 0$ in this case, the right side of the above equation is less than 1. In other words, if $p_1 \geq \alpha p_j$, then we can determine that $\mu_j=0$. On the other hand, by setting the value of $j$ in equation (\ref{KL=}) and multiplying these expressions together, we can obtain the following relationship:
    \begin{equation}
        \alpha^{|A|}\exp(-\alpha \sum_{j \in A} \mu_j)\prod_{j \in A} p_j = p_1^{|A|} \exp(|A|\sum_{k \in A} \mu_k).
    \end{equation}
    Here, $A$ is defined as the set of indices $k$ such that $\mu_k>0$, i.e., $A := \{k|\mu_k>0\}$. $|A|$ denotes the cardinality of set $A$.
    Then, we can derive:
    \begin{equation}
        \exp(\sum_{k \in A} \mu_k)= (p_1^{-|A|} (\prod_{j\in A} \alpha p_j))^{\frac{1}{|A|+\alpha}}.
    \end{equation}
    By using the above equation, we can express $q_1$ and $q_j$ without using $\mu_j$:
    \begin{align}
        q_1 = \alpha q_j
        = \exp{(-1-\lambda)} (p_1^{\alpha} (\prod_{j\in A} \alpha p_j))^{\frac{1}{|A|+\alpha}}.  \label{KL_xj_}
    \end{align}
    \item \textbf{Case 2:} $\mu_j = 0$.
    
    If $\mu_j = 0$, $q_j$ can be easily expressed without using $\mu_j$:
        \begin{equation} 
            q_j=p_j \exp{(-1-\lambda)}. \label{KL_xj=_}
        \end{equation}
    Then, by using the equations (\ref{KL=}), (\ref{KL_xj_}) and (\ref{KL_xj=_}), we can also express the objective function without using $\mu_j$:
    \begin{align} \label{KL_obj}
    \sum_{i=1}^{N} q_i \log \frac{q_i}{p_i} = - (1+\lambda).
    \end{align}
    Finally, by applying the normalization condition $\sum_{i=1}^N q_i =1$, we can solve for the value of $\lambda$ using the following equation:
    \begin{equation} \label{KL_lambda}
        \exp(1+\lambda) = (1+\frac{|A|}{\alpha})(p_1^{\alpha} (\prod_{j\in A}\alpha p_j))^{\frac{1}{|A|+\alpha}} + \sum_{j \not\in A\cup \{1\} }p_j. 
    \end{equation}
\end{itemize}

\subsubsection{Determine Active Constraints}
To determine the active constraints, we have the following proposition, which directly leads to the formulation of Algorithm \ref{alg:PCE}. 

\begin{proposition} Consider the optimization problem in (\ref{op}).
 Let $A$ be the set of indices $k$ such that $\mu_k > 0$, i.e., $A:=\{k|\mu_k>0\}$. Define $B$ as the set $\{p_j|\alpha p_j>p_1\}$.  Sort the elements of $B$ in descending order as:
\begin{equation} 
p_{j_1} \geq p_{j_2} \geq \ldots \geq p_{j_{|B|}}.
\end{equation}
Note that duplicate elements in $B$ are not removed.
Then, the following conclusions holds:
\begin{enumerate}
    \item  The index $j_1$, which corresponds to the largest element in $B$, must belong to $A$, i.e., $j_1 \in A$.
    \item  For any $C=\{j_1,\ldots,j_{t-1}\} \subseteq A$, if the inequality
        \begin{equation}
             (p_1^{\alpha} (\prod_{j\in C}\alpha p_j))^{\frac{1}{|C|+\alpha}} < \alpha p_{j_t}
        \end{equation}
        holds, then $j_t \in A$. Otherwise, for all $s \in \{t,t+1, \ldots, |B|\}$, we have $j_s \not\in A$. That is, $A=C$.
\end{enumerate}
\end{proposition}

\begin{proof}

Firstly, we can prove that $j_1 \in A$. To do so, assume for the sake of contradiction that $j_1 \notin A$. By using equations  (\ref{KL_xj_}) and (\ref{KL_xj=_}), we obtain the following expression for $q_1$:
\begin{align}
    q_1 &= \exp{(-1-\lambda)} (p_1^{\alpha} (\prod_{j\in A} \alpha p_j))^{\frac{1}{|A|+\alpha}} \nonumber\\
    &<  \exp{(-1-\lambda)} ((\alpha p_{j_1})^{\alpha} (\prod_{j\in A}\alpha p_{j_1}))^{\frac{1}{|A|+\alpha}} \nonumber\\
    &= \alpha \exp{(-1-\lambda)} p_{j_1} = \alpha q_{j_1}.
\end{align}
Here, the first equality follows from the equation (\ref{KL_xj_}), and the inequality in the second-to-last line is based on the definition of B, in which $p_1<\alpha p_{j_1}$. The first equality in  the last line is simply an identity transformation, and the last equation uses the equation (\ref{KL_xj=_}) under the assumption that $j_1 \notin A$.
The above analysis shows that $q_1 < \alpha q_{j_1}$, which clearly violates the inequality constraints of the original problem. Thus, our assumption that $j_1 \notin A $ must be false.

Secondly, we consider the case where $C = \{j_1, \ldots, j_{t-1}\} \subseteq A$, and we assume that the inequality
\begin{equation}\label{cond:in}
     (p_1^{\alpha} (\prod_{j\in C}\alpha p_j))^{\frac{1}{|C|+\alpha}} < \alpha p_{j_t}
\end{equation}
holds. We now want to prove that $j_t \in A$. Suppose, for contradiction, that $j_t \notin A$. In this case, we proceed as follows. Using the equations (\ref{KL_xj_}) and (\ref{KL_xj=_}), we can express $q_1$ as:
\begin{align}
    q_1 &= \exp{(-1-\lambda)} (p_1^{\alpha} (\prod_{j\in A}\alpha p_j))^{\frac{1}{|A|+\alpha}} \nonumber\\
    &= \exp{(-1-\lambda)} (p_1^{\alpha} (\prod_{j\in C}\alpha p_j) (\prod_{j\in A-C}\alpha p_j))^{\frac{1}{|A|+\alpha}} \nonumber\\
    &< \exp{(-1-\lambda)} (p_1^{\alpha} (\prod_{j\in C}\alpha p_j) (\prod_{j\in A-C}\alpha p_{j_t}))^{\frac{1}{|A|+\alpha}} \nonumber\\
    &<  \exp{(-1-\lambda)} ((\alpha p_{j_t})^{|C|+\alpha} (\prod_{j\in A-C}\alpha p_{j_t}))^{\frac{1}{|A|+\alpha}} \nonumber\\ 
    &= \alpha \exp{(-1-\lambda)} p_{j_t}= \alpha q_{j_t}.
\end{align}
The first equality uses equation (\ref{KL_xj_}). The first inequality follows from the fact that $p_{j_t}>p_{j}$ for any $j\in A-C$, which is because the elements of $B$ have been sorted in descending order. The second inequality uses the condition stated in (\ref{cond:in}). The last equality follows from equation (\ref{KL_xj=_}) under the assumption that $j_t \notin A$.
The above analysis shows that  $q_1 < \alpha q_{j_t}$, 
which, again, violates the inequality constraints of the original problem. Hence, our assumption that $j_t \notin A $ is false, and it holds that $j_t \in A $.

Finally, we address the case where the inequality (\ref{cond:in}) does not holds. In this case, for all $s \in \{t,t+1, \ldots, |B|\}$, we assert that $j_s \not\in A$. We will now prove this by considering two cases.
In the first case, we assume that there exists some $j_s \in A$ such that 
$p_1^{\alpha} (\prod_{j\in C}\alpha p_j))^{\frac{1}{|C|+\alpha}} > \alpha p_{j_s}$.
If $j_s\in A$, we then have:
\begin{align} 
&\quad\sum_{i=1}^{N} q_i \log \frac{q_i}{p_i} = - (1+\lambda) \nonumber \\
&= -\log\left[ (1+\frac{|C|+1}{\alpha})((p_1^{\alpha} (\prod_{j\in C \cup \{j_s\}}\alpha p_j))^{\frac{1}{|C|+1+\alpha}} \right.\nonumber \\
&\left.\quad + \sum_{j \not\in C\cup \{1,j_s\} }p_j\right].
\end{align}
The equations (\ref{KL_obj}) and (\ref{KL_lambda}) have been applied above.
However, if $j_s\not\in A$, we have
\begin{align} 
     &\quad\sum_{i=1}^{N} q_i \log \frac{q_i}{p_i} = - (1+\lambda) \nonumber \\
    &= -\log \left[ (1+\frac{|C|}{\alpha})(p_1^{\alpha} (\prod_{j\in C}\alpha p_j))^{\frac{1}{|C|+\alpha}} + \sum_{j \not\in C\cup \{1\} }p_j\right].
\end{align}
Now we consider the function of $p_{j_s}$:
\begin{align}
    f(p_{j_s})&= (1+\frac{|C|}{\alpha})(p_1^{\alpha} (\prod_{j\in C}\alpha p_j))^{\frac{1}{|C|+\alpha}} + p_{j_s} \nonumber \\
    & \quad- (1+\frac{|C|+1}{\alpha})(p_1^{\alpha} (\alpha p_{j_s}) (\prod_{j\in C}\alpha p_j))^{\frac{1}{|C|+1+\alpha}}.
\end{align}
Then, the first derivative of $f(p_{j_s})$  with respect to $p_{j_s}$ is given by:
\begin{equation}
    f'(p_{j_s}) = 1-\frac{1}{\alpha}(\alpha p_1^{\alpha} (\prod_{j\in C}\alpha p_j))^{\frac{1}{|C|+1+\alpha}} p_{j_s}^{\frac{1}{|C|+1+\alpha}-1}.
\end{equation}
Obviously, $f'(p_{j_s})$ increases gradually as $p_{j_s}$ increases, and $f'(\frac{1}{\alpha}( p_1^{\alpha} (\prod_{j\in C}\alpha p_j))^{\frac{1}{|C|+\alpha}})=0$. 
Since $(p_1^{\alpha} (\prod_{j\in C}\alpha p_j))^{\frac{1}{|C|+\alpha}} > \alpha p_{j_s}$,
we can get
\begin{equation} \label{KL_optimal}
    f(p_{j_s}) > f(\frac{1}{\alpha}( p_1^{\alpha} (\prod_{j\in C} \alpha p_j))^{\frac{1}{|C|+\alpha}}) = 0.
\end{equation}
From equation  (\ref{KL_optimal}), we can observe that the value of the objective function is smaller if $j_s\not\in A$.
In the second case, if there exists $j_s$ such that $(p_1^{\alpha} (\prod_{j\in C}\alpha p_j))^{\frac{1}{|C|+\alpha}} = \alpha p_{j_s}$ and $j_s\in A$, we have
\begin{align}
q_{j_s} &= \frac{q_1}{\alpha} \nonumber \\
&= \frac{1}{\alpha}\exp{(-1-\lambda)} 
(p_1^{\alpha} (\alpha p_{j_s}) (\prod_{j\in C}\alpha p_j))^{\frac{1}{|C|+1+\alpha}} \nonumber \\
&= p_{j_s}\exp(-1-\lambda),
\end{align}
which leads to the conclusion that $\mu_{j_s}$ = 0 according to the equation (\ref{KL_xj}), contradicting the definition of the set $A$. Hence, no such $j_s$ can exist.
In conclusion, for all $s \in \{t,t+1, \ldots, |B|\}$, it holds that $j_s \not\in A$.

\end{proof}

At this time, we have successfully identified the active set $A$. Subsequently, we can obtain the optimal solution by applying equations  (\ref{KL_xj_}), (\ref{KL_xj=_}), and (\ref{KL_lambda}), which leads to the formulation of Algorithm \ref{alg:PCE}.

\subsection{Time Efficiency}

Although the proof may be somewhat intricate, it is worth emphasizing that Algorithm \ref{alg:PCE} exhibits significantly higher efficiency in comparison to conventional convex programming tools such as MOSEK. The table below presents the average time taken by the classical convex programming tool (MOSEK) and the proposed Algorithm \ref{alg:PCE} over 1000 runs. It is evident that for $N=100$, Algorithm \ref{alg:PCE} consumes only around one-tenth of the time required by traditional convex programming tools.

\setlength{\tabcolsep}{12.5pt}
\begin{table}[H]
  \caption{\small {Comparisons about average training time required by Algorithm \ref{alg:PCE} and the MOSEK tool.} }
  \centering
  \scalebox{0.8}{\begin{tabular}{c|c|c|c}
    \toprule
    & MOSEK & Algo. \ref{alg:PCE} (Ours) & Speedup ratio \\
    \midrule
    Time (Avg.) & 21.93 ms & 2.16 ms & 10.15 \\
    \bottomrule
  \end{tabular}}
\end{table}

\subsection{Further Discussion}
By examining Algorithm \ref{alg:PCE}, we can find that if $\alpha$ is sufficiently large such that $|A|=N-1$, the optimal soft labels degenerate into smoothed labels with a label smoothing factor of $s=N/(\alpha-1+N)$. Furthermore, as $\alpha$ increases towards infinity, the optimal soft labels converge towards one-hot labels. Then, the PCE loss degenerate into the traditional cross entropy.

\section{Full Results with ResNet-50}\label{section:3}
In this section, we demonstrate the full results of our experiments that were carried out on five well-known benchmark datasets, including VLCS, PACS, OfficeHome, TerraIncognita and DomainNet. These experiments were conducted using ResNet-50, which was pre-trained on ImageNet.

\subsection{VLCS}
\setlength{\tabcolsep}{22pt}
\begin{table}[H]
  \caption{\small {Out-of-domain accuracies (\%) on each domain of VLCS and their average.} }
  \centering
  \scalebox{0.5}{\begin{tabular}{l|cccc|c}
    \toprule
     {Algorithms} &  {C} &  {L} &  {S} &  {V} &  {Avg.} \\
    \midrule
    ERM                  & 97.7 \scriptsize{$\pm$ 0.4}       & 64.3 \scriptsize{$\pm$ 0.9}       & 73.4 \scriptsize{$\pm$ 0.5}       & 74.6 \scriptsize{$\pm$ 1.3}       & 77.5                 \\
    IRM                  & 98.6 \scriptsize{$\pm$ 0.1}       & 64.9 \scriptsize{$\pm$ 0.9}       & 73.4 \scriptsize{$\pm$ 0.6}       & 77.3 \scriptsize{$\pm$ 0.9}       & 78.5                 \\
    GroupDRO             & 97.3 \scriptsize{$\pm$ 0.3}       & 63.4 \scriptsize{$\pm$ 0.9}       & 69.5 \scriptsize{$\pm$ 0.8}       & 76.7 \scriptsize{$\pm$ 0.7}       & 76.7                 \\
    Mixup                & 98.3 \scriptsize{$\pm$ 0.6}       & 64.8 \scriptsize{$\pm$ 1.0}       & 72.1 \scriptsize{$\pm$ 0.5}       & 74.3 \scriptsize{$\pm$ 0.8}       & 77.4                 \\
    MLDG                 & 97.4 \scriptsize{$\pm$ 0.2}       & 65.2 \scriptsize{$\pm$ 0.7}       & 71.0 \scriptsize{$\pm$ 1.4}       & 75.3 \scriptsize{$\pm$ 1.0}       & 77.2                 \\
    CORAL                & 98.3 \scriptsize{$\pm$ 0.1}       & 66.1 \scriptsize{$\pm$ 1.2}       & 73.4 \scriptsize{$\pm$ 0.3}       & 77.5 \scriptsize{$\pm$ 1.2}       & 78.8                 \\
    MMD                  & 97.7 \scriptsize{$\pm$ 0.1}       & 64.0 \scriptsize{$\pm$ 1.1}       & 72.8 \scriptsize{$\pm$ 0.2}       & 75.3 \scriptsize{$\pm$ 3.3}       & 77.5                 \\
    DANN                 & 99.0 \scriptsize{$\pm$ 0.3}       & 65.1 \scriptsize{$\pm$ 1.4}       & 73.1 \scriptsize{$\pm$ 0.3}       & 77.2 \scriptsize{$\pm$ 0.6}       & 78.6                 \\
    CDANN                & 97.1 \scriptsize{$\pm$ 0.3}       & 65.1 \scriptsize{$\pm$ 1.2}       & 70.7 \scriptsize{$\pm$ 0.8}       & 77.1 \scriptsize{$\pm$ 1.5}       & 77.5                 \\
    MTL                  & 97.8 \scriptsize{$\pm$ 0.4}       & 64.3 \scriptsize{$\pm$ 0.3}       & 71.5 \scriptsize{$\pm$ 0.7}       & 75.3 \scriptsize{$\pm$ 1.7}       & 77.2                 \\
    SagNet               & 97.9 \scriptsize{$\pm$ 0.4}       & 64.5 \scriptsize{$\pm$ 0.5}       & 71.4 \scriptsize{$\pm$ 1.3}       & 77.5 \scriptsize{$\pm$ 0.5}       & 77.8                 \\
    ARM                  & 98.7 \scriptsize{$\pm$ 0.2}       & 63.6 \scriptsize{$\pm$ 0.7}       & 71.3 \scriptsize{$\pm$ 1.2}       & 76.7 \scriptsize{$\pm$ 0.6}       & 77.6                 \\
    VREx                 & 98.4 \scriptsize{$\pm$ 0.3}       & 64.4 \scriptsize{$\pm$ 1.4}       & 74.1 \scriptsize{$\pm$ 0.4}       & 76.2 \scriptsize{$\pm$ 1.3}       & 78.3                 \\
    RSC                  & 97.9 \scriptsize{$\pm$ 0.1}       & 62.5 \scriptsize{$\pm$ 0.7}       & 72.3 \scriptsize{$\pm$ 1.2}       & 75.6 \scriptsize{$\pm$ 0.8}       & 77.1                 \\
    \midrule
    SFT (Ours)            & 99.5 \scriptsize{$\pm$ 0.1}       & 66.2 \scriptsize{$\pm$ 0.2}       & 74.8 \scriptsize{$\pm$ 0.5}       & 78.7 \scriptsize{$\pm$ 0.4}       & 79.8                 \\
    
    \bottomrule
  \end{tabular}}
\end{table}

\subsection{PACS}
\setlength{\tabcolsep}{22pt}
\begin{table}[H]
  \caption{\small {Out-of-domain accuracies (\%) on each domain of PACS and their average.} }
  \centering
  \scalebox{0.5}{\begin{tabular}{l|cccc|c}
    \toprule
     {Algorithms} &  {A} &  {C} &  {P} &  {S} &  {Avg.} \\
    \midrule
    ERM                  & 84.7 \scriptsize{$\pm$ 0.4}       & 80.8 \scriptsize{$\pm$ 0.6}       & 97.2 \scriptsize{$\pm$ 0.3}       & 79.3 \scriptsize{$\pm$ 1.0}       & 85.5                 \\
    IRM                  & 84.8 \scriptsize{$\pm$ 1.3}       & 76.4 \scriptsize{$\pm$ 1.1}       & 96.7 \scriptsize{$\pm$ 0.6}       & 76.1 \scriptsize{$\pm$ 1.0}       & 83.5                 \\
    GroupDRO             & 83.5 \scriptsize{$\pm$ 0.9}       & 79.1 \scriptsize{$\pm$ 0.6}       & 96.7 \scriptsize{$\pm$ 0.3}       & 78.3 \scriptsize{$\pm$ 2.0}       & 84.4                 \\
    Mixup                & 86.1 \scriptsize{$\pm$ 0.5}       & 78.9 \scriptsize{$\pm$ 0.8}       & 97.6 \scriptsize{$\pm$ 0.1}       & 75.8 \scriptsize{$\pm$ 1.8}       & 84.6                 \\
    MLDG                 & 85.5 \scriptsize{$\pm$ 1.4}       & 80.1 \scriptsize{$\pm$ 1.7}       & 97.4 \scriptsize{$\pm$ 0.3}       & 76.6 \scriptsize{$\pm$ 1.1}       & 84.9                 \\
    CORAL                & 88.3 \scriptsize{$\pm$ 0.2}       & 80.0 \scriptsize{$\pm$ 0.5}       & 97.5 \scriptsize{$\pm$ 0.3}       & 78.8 \scriptsize{$\pm$ 1.3}       & 86.2                 \\
    MMD                  & 86.1 \scriptsize{$\pm$ 1.4}       & 79.4 \scriptsize{$\pm$ 0.9}       & 96.6 \scriptsize{$\pm$ 0.2}       & 76.5 \scriptsize{$\pm$ 0.5}       & 84.6                 \\
    DANN                 & 86.4 \scriptsize{$\pm$ 0.8}       & 77.4 \scriptsize{$\pm$ 0.8}       & 97.3 \scriptsize{$\pm$ 0.4}       & 73.5 \scriptsize{$\pm$ 2.3}       & 83.6                 \\
    CDANN                & 84.6 \scriptsize{$\pm$ 1.8}       & 75.5 \scriptsize{$\pm$ 0.9}       & 96.8 \scriptsize{$\pm$ 0.3}       & 73.5 \scriptsize{$\pm$ 0.6}       & 82.6                 \\
    MTL                  & 87.5 \scriptsize{$\pm$ 0.8}       & 77.1 \scriptsize{$\pm$ 0.5}       & 96.4 \scriptsize{$\pm$ 0.8}       & 77.3 \scriptsize{$\pm$ 1.8}       & 84.6                 \\
    SagNet               & 87.4 \scriptsize{$\pm$ 1.0}       & 80.7 \scriptsize{$\pm$ 0.6}       & 97.1 \scriptsize{$\pm$ 0.1}       & 80.0 \scriptsize{$\pm$ 0.4}       & 86.3                 \\
    ARM                  & 86.8 \scriptsize{$\pm$ 0.6}       & 76.8 \scriptsize{$\pm$ 0.5}       & 97.4 \scriptsize{$\pm$ 0.3}       & 79.3 \scriptsize{$\pm$ 1.2}       & 85.1                 \\
    VREx                 & 86.0 \scriptsize{$\pm$ 1.6}       & 79.1 \scriptsize{$\pm$ 0.6}       & 96.9 \scriptsize{$\pm$ 0.5}       & 77.7 \scriptsize{$\pm$ 1.7}       & 84.9                 \\
    RSC                  & 85.4 \scriptsize{$\pm$ 0.8}       & 79.7 \scriptsize{$\pm$ 1.8}       & 97.6 \scriptsize{$\pm$ 0.3}       & 78.2 \scriptsize{$\pm$ 1.2}       & 85.2                 \\ 
    \midrule
    SFT (Ours)             & 90.1 \scriptsize{$\pm$ 0.3}       & 80.3 \scriptsize{$\pm$ 1.0}       & 98.6 \scriptsize{$\pm$ 0.2}       & 84.3 \scriptsize{$\pm$ 1.4}       & 88.3                 \\ 
    
    \bottomrule
  \end{tabular}}
\end{table}
 
\subsection{OfficeHome}
\setlength{\tabcolsep}{22pt}
\begin{table}[H]
  \caption{\small {Out-of-domain accuracies (\%) on each domain of OfficeHome and their average.} }
  \centering
  \scalebox{0.5}{\begin{tabular}{l|cccc|c}
    \toprule
     {Algorithms} &  {A} &  {C} &  {P} &  {R} &  {Avg.} \\
    \midrule
    ERM                  & 61.3 \scriptsize{$\pm$ 0.7}       & 52.4 \scriptsize{$\pm$ 0.3}       & 75.8 \scriptsize{$\pm$ 0.1}       & 76.6 \scriptsize{$\pm$ 0.3}       & 66.5                 \\
    IRM                  & 58.9 \scriptsize{$\pm$ 2.3}       & 52.2 \scriptsize{$\pm$ 1.6}       & 72.1 \scriptsize{$\pm$ 2.9}       & 74.0 \scriptsize{$\pm$ 2.5}       & 64.3                 \\
    GroupDRO             & 60.4 \scriptsize{$\pm$ 0.7}       & 52.7 \scriptsize{$\pm$ 1.0}       & 75.0 \scriptsize{$\pm$ 0.7}       & 76.0 \scriptsize{$\pm$ 0.7}       & 66.0                 \\
    Mixup                & 62.4 \scriptsize{$\pm$ 0.8}       & 54.8 \scriptsize{$\pm$ 0.6}       & 76.9 \scriptsize{$\pm$ 0.3}       & 78.3 \scriptsize{$\pm$ 0.2}       & 68.1                 \\
    MLDG                 & 61.5 \scriptsize{$\pm$ 0.9}       & 53.2 \scriptsize{$\pm$ 0.6}       & 75.0 \scriptsize{$\pm$ 1.2}       & 77.5 \scriptsize{$\pm$ 0.4}       & 66.8                 \\
    CORAL                & 65.3 \scriptsize{$\pm$ 0.4}       & 54.4 \scriptsize{$\pm$ 0.5}       & 76.5 \scriptsize{$\pm$ 0.1}       & 78.4 \scriptsize{$\pm$ 0.5}       & 68.7                 \\
    MMD                  & 60.4 \scriptsize{$\pm$ 0.2}       & 53.3 \scriptsize{$\pm$ 0.3}       & 74.3 \scriptsize{$\pm$ 0.1}       & 77.4 \scriptsize{$\pm$ 0.6}       & 66.3                 \\
    DANN                 & 59.9 \scriptsize{$\pm$ 1.3}       & 53.0 \scriptsize{$\pm$ 0.3}       & 73.6 \scriptsize{$\pm$ 0.7}       & 76.9 \scriptsize{$\pm$ 0.5}       & 65.9                 \\
    CDANN                & 61.5 \scriptsize{$\pm$ 1.4}       & 50.4 \scriptsize{$\pm$ 2.4}       & 74.4 \scriptsize{$\pm$ 0.9}       & 76.6 \scriptsize{$\pm$ 0.8}       & 65.8                 \\
    MTL                  & 61.5 \scriptsize{$\pm$ 0.7}       & 52.4 \scriptsize{$\pm$ 0.6}       & 74.9 \scriptsize{$\pm$ 0.4}       & 76.8 \scriptsize{$\pm$ 0.4}       & 66.4                 \\
    SagNet               & 63.4 \scriptsize{$\pm$ 0.2}       & 54.8 \scriptsize{$\pm$ 0.4}       & 75.8 \scriptsize{$\pm$ 0.4}       & 78.3 \scriptsize{$\pm$ 0.3}       & 68.1                 \\
    ARM                  & 58.9 \scriptsize{$\pm$ 0.8}       & 51.0 \scriptsize{$\pm$ 0.5}       & 74.1 \scriptsize{$\pm$ 0.1}       & 75.2 \scriptsize{$\pm$ 0.3}       & 64.8                 \\
    VREx                 & 60.7 \scriptsize{$\pm$ 0.9}       & 53.0 \scriptsize{$\pm$ 0.9}       & 75.3 \scriptsize{$\pm$ 0.1}       & 76.6 \scriptsize{$\pm$ 0.5}       & 66.4                 \\
    RSC                  & 60.7 \scriptsize{$\pm$ 1.4}       & 51.4 \scriptsize{$\pm$ 0.3}       & 74.8 \scriptsize{$\pm$ 1.1}       & 75.1 \scriptsize{$\pm$ 1.3}       & 65.5                 \\
    \midrule
    SFT (Ours)              & 65.8 \scriptsize{$\pm$ 0.3}       & 58.8 \scriptsize{$\pm$ 0.3}       & 78.3 \scriptsize{$\pm$ 0.4}       & 80.6 \scriptsize{$\pm$ 0.2}       & 70.9                 \\
    \bottomrule
  \end{tabular}}
\end{table}

\subsection{TerraIncognita}
\setlength{\tabcolsep}{20pt}
\begin{table}[H]
  \caption{\small {Out-of-domain accuracies (\%) on each domain of TerraIncognita and their average.} }
  \centering
  \scalebox{0.5}{\begin{tabular}{l|cccc|c}
    \toprule
     {Algorithms} &  {L100} &  {L38} &  {L43} &  {L46} &  {Avg.} \\
    \midrule
    ERM                  & 49.8 \scriptsize{$\pm$ 4.4}       & 42.1 \scriptsize{$\pm$ 1.4}       & 56.9 \scriptsize{$\pm$ 1.8}       & 35.7 \scriptsize{$\pm$ 3.9}       & 46.1                 \\
    IRM                  & 54.6 \scriptsize{$\pm$ 1.3}       & 39.8 \scriptsize{$\pm$ 1.9}       & 56.2 \scriptsize{$\pm$ 1.8}       & 39.6 \scriptsize{$\pm$ 0.8}       & 47.6                 \\
    GroupDRO             & 41.2 \scriptsize{$\pm$ 0.7}       & 38.6 \scriptsize{$\pm$ 2.1}       & 56.7 \scriptsize{$\pm$ 0.9}       & 36.4 \scriptsize{$\pm$ 2.1}       & 43.2                 \\
    Mixup                & 59.6 \scriptsize{$\pm$ 2.0}       & 42.2 \scriptsize{$\pm$ 1.4}       & 55.9 \scriptsize{$\pm$ 0.8}       & 33.9 \scriptsize{$\pm$ 1.4}       & 47.9                 \\
    MLDG                 & 54.2 \scriptsize{$\pm$ 3.0}       & 44.3 \scriptsize{$\pm$ 1.1}       & 55.6 \scriptsize{$\pm$ 0.3}       & 36.9 \scriptsize{$\pm$ 2.2}       & 47.7                 \\
    CORAL                & 51.6 \scriptsize{$\pm$ 2.4}       & 42.2 \scriptsize{$\pm$ 1.0}       & 57.0 \scriptsize{$\pm$ 1.0}       & 39.8 \scriptsize{$\pm$ 2.9}       & 47.6                 \\
    MMD                  & 41.9 \scriptsize{$\pm$ 3.0}       & 34.8 \scriptsize{$\pm$ 1.0}       & 57.0 \scriptsize{$\pm$ 1.9}       & 35.2 \scriptsize{$\pm$ 1.8}       & 42.2                 \\
    DANN                 & 51.1 \scriptsize{$\pm$ 3.5}       & 40.6 \scriptsize{$\pm$ 0.6}       & 57.4 \scriptsize{$\pm$ 0.5}       & 37.7 \scriptsize{$\pm$ 1.8}       & 46.7                 \\
    CDANN                & 47.0 \scriptsize{$\pm$ 1.9}       & 41.3 \scriptsize{$\pm$ 4.8}       & 54.9 \scriptsize{$\pm$ 1.7}       & 39.8 \scriptsize{$\pm$ 2.3}       & 45.8                 \\
    MTL                  & 49.3 \scriptsize{$\pm$ 1.2}       & 39.6 \scriptsize{$\pm$ 6.3}       & 55.6 \scriptsize{$\pm$ 1.1}       & 37.8 \scriptsize{$\pm$ 0.8}       & 45.6                 \\
    SagNet               & 53.0 \scriptsize{$\pm$ 2.9}       & 43.0 \scriptsize{$\pm$ 2.5}       & 57.9 \scriptsize{$\pm$ 0.6}       & 40.4 \scriptsize{$\pm$ 1.3}       & 48.6                 \\
    ARM                  & 49.3 \scriptsize{$\pm$ 0.7}       & 38.3 \scriptsize{$\pm$ 2.4}       & 55.8 \scriptsize{$\pm$ 0.8}       & 38.7 \scriptsize{$\pm$ 1.3}       & 45.5                 \\
    VREx                 & 48.2 \scriptsize{$\pm$ 4.3}       & 41.7 \scriptsize{$\pm$ 1.3}       & 56.8 \scriptsize{$\pm$ 0.8}       & 38.7 \scriptsize{$\pm$ 3.1}       & 46.4                 \\
    RSC                  & 50.2 \scriptsize{$\pm$ 2.2}       & 39.2 \scriptsize{$\pm$ 1.4}       & 56.3 \scriptsize{$\pm$ 1.4}       & 40.8 \scriptsize{$\pm$ 0.6}       & 46.6                 \\
     
    \midrule
    SFT (Ours)      & 57.5 \scriptsize{$\pm$ 0.4}       & 44.6 \scriptsize{$\pm$ 1.4}       & 59.6 \scriptsize{$\pm$ 0.5}       & 41.0 \scriptsize{$\pm$ 1.0}       & 50.7                 \\
    
    \bottomrule
  \end{tabular}}
\end{table}

\subsection{DomainNet}
\setlength{\tabcolsep}{13pt}
\begin{table}[H]
  \caption{\small {Out-of-domain accuracies (\%) on each domain of TerraIncognita and their average.} }
  \centering
  \scalebox{0.45}{\begin{tabular}{l|cccccc|c}
    \toprule
     {Algorithm}   &  {clip}        &  {info}        &  {paint}       &  {quick}       &  {real}        &  {sketch}      &  {Avg.}         \\
    \midrule
    ERM                  & 58.1 \scriptsize{$\pm$ 0.3}       & 18.8 \scriptsize{$\pm$ 0.3}       & 46.7 \scriptsize{$\pm$ 0.3}       & 12.2 \scriptsize{$\pm$ 0.4}       & 59.6 \scriptsize{$\pm$ 0.1}       & 49.8 \scriptsize{$\pm$ 0.4}       & 40.9                 \\
    IRM                  & 48.5 \scriptsize{$\pm$ 2.8}       & 15.0 \scriptsize{$\pm$ 1.5}       & 38.3 \scriptsize{$\pm$ 4.3}       & 10.9 \scriptsize{$\pm$ 0.5}       & 48.2 \scriptsize{$\pm$ 5.2}       & 42.3 \scriptsize{$\pm$ 3.1}       & 33.9                 \\
    GroupDRO             & 47.2 \scriptsize{$\pm$ 0.5}       & 17.5 \scriptsize{$\pm$ 0.4}       & 33.8 \scriptsize{$\pm$ 0.5}       & 9.3 \scriptsize{$\pm$ 0.3}        & 51.6 \scriptsize{$\pm$ 0.4}       & 40.1 \scriptsize{$\pm$ 0.6}       & 33.3                 \\
    Mixup                & 55.7 \scriptsize{$\pm$ 0.3}       & 18.5 \scriptsize{$\pm$ 0.5}       & 44.3 \scriptsize{$\pm$ 0.5}       & 12.5 \scriptsize{$\pm$ 0.4}       & 55.8 \scriptsize{$\pm$ 0.3}       & 48.2 \scriptsize{$\pm$ 0.5}       & 39.2                 \\
    MLDG                 & 59.1 \scriptsize{$\pm$ 0.2}       & 19.1 \scriptsize{$\pm$ 0.3}       & 45.8 \scriptsize{$\pm$ 0.7}       & 13.4 \scriptsize{$\pm$ 0.3}       & 59.6 \scriptsize{$\pm$ 0.2}       & 50.2 \scriptsize{$\pm$ 0.4}       & 41.2                 \\
    CORAL                & 59.2 \scriptsize{$\pm$ 0.1}       & 19.7 \scriptsize{$\pm$ 0.2}       & 46.6 \scriptsize{$\pm$ 0.3}       & 13.4 \scriptsize{$\pm$ 0.4}       & 59.8 \scriptsize{$\pm$ 0.2}       & 50.1 \scriptsize{$\pm$ 0.6}       & 41.5                 \\
    MMD                  & 32.1 \scriptsize{$\pm$ 13.3}      & 11.0 \scriptsize{$\pm$ 4.6}       & 26.8 \scriptsize{$\pm$ 11.3}      & 8.7 \scriptsize{$\pm$ 2.1}        & 32.7 \scriptsize{$\pm$ 13.8}      & 28.9 \scriptsize{$\pm$ 11.9}      & 23.4                 \\
    DANN                 & 53.1 \scriptsize{$\pm$ 0.2}       & 18.3 \scriptsize{$\pm$ 0.1}       & 44.2 \scriptsize{$\pm$ 0.7}       & 11.8 \scriptsize{$\pm$ 0.1}       & 55.5 \scriptsize{$\pm$ 0.4}       & 46.8 \scriptsize{$\pm$ 0.6}       & 38.3                 \\
    CDANN                & 54.6 \scriptsize{$\pm$ 0.4}       & 17.3 \scriptsize{$\pm$ 0.1}       & 43.7 \scriptsize{$\pm$ 0.9}       & 12.1 \scriptsize{$\pm$ 0.7}       & 56.2 \scriptsize{$\pm$ 0.4}       & 45.9 \scriptsize{$\pm$ 0.5}       & 38.3                 \\
    MTL                  & 57.9 \scriptsize{$\pm$ 0.5}       & 18.5 \scriptsize{$\pm$ 0.4}       & 46.0 \scriptsize{$\pm$ 0.1}       & 12.5 \scriptsize{$\pm$ 0.1}       & 59.5 \scriptsize{$\pm$ 0.3}       & 49.2 \scriptsize{$\pm$ 0.1}       & 40.6                 \\
    SagNet               & 57.7 \scriptsize{$\pm$ 0.3}       & 19.0 \scriptsize{$\pm$ 0.2}       & 45.3 \scriptsize{$\pm$ 0.3}       & 12.7 \scriptsize{$\pm$ 0.5}       & 58.1 \scriptsize{$\pm$ 0.5}       & 48.8 \scriptsize{$\pm$ 0.2}       & 40.3                 \\
    ARM                  & 49.7 \scriptsize{$\pm$ 0.3}       & 16.3 \scriptsize{$\pm$ 0.5}       & 40.9 \scriptsize{$\pm$ 1.1}       & 9.4 \scriptsize{$\pm$ 0.1}        & 53.4 \scriptsize{$\pm$ 0.4}       & 43.5 \scriptsize{$\pm$ 0.4}       & 35.5                 \\
    VREx                 & 47.3 \scriptsize{$\pm$ 3.5}       & 16.0 \scriptsize{$\pm$ 1.5}       & 35.8 \scriptsize{$\pm$ 4.6}       & 10.9 \scriptsize{$\pm$ 0.3}       & 49.6 \scriptsize{$\pm$ 4.9}       & 42.0 \scriptsize{$\pm$ 3.0}       & 33.6                 \\
    RSC                  & 55.0 \scriptsize{$\pm$ 1.2}       & 18.3 \scriptsize{$\pm$ 0.5}       & 44.4 \scriptsize{$\pm$ 0.6}       & 12.2 \scriptsize{$\pm$ 0.2}       & 55.7 \scriptsize{$\pm$ 0.7}       & 47.8 \scriptsize{$\pm$ 0.9}       & 38.9                 \\
    \midrule
    SFT (Ours)      & 64.9 \scriptsize{$\pm$ 0.0}       & 22.0 \scriptsize{$\pm$ 0.3}       & 52.5 \scriptsize{$\pm$ 0.1}       & 16.3 \scriptsize{$\pm$ 0.3}       & 64.4 \scriptsize{$\pm$ 0.1}       & 55.6 \scriptsize{$\pm$ 0.4}       & 46.0            \\
    
    \bottomrule
  \end{tabular}}
\end{table}

\section{Full Results with ViT-B/16 and ViT-L/14}\label{section:4}
In this section, we show the full results of visual prompt tuning with the pre-trained large-scale vision transformers (including ViT-B/16 and ViT-L/14). 
 
\subsection{VLCS}
\setlength{\tabcolsep}{18pt}
\begin{table}[H]
  \caption{\small {Out-of-domain accuracies (\%) on each domain of VLCS and their average.}}
  \centering
  \scalebox{0.5}{\begin{tabular}{c|l|cccc|c}
    \toprule
     {Backbone} &  {Algorithms} &  {C} &  {L} &  {S} &  {V} &  {Avg}. \\
    \midrule
    \multirow{9}{*}{ViT-B/16} & ERM & 95.9 & 66.2 & 81.8 & 79.7 & 80.9 \\
    & IRM & 96.5 & 67.7 &  {85.0} & 78.7 & 81.9 \\
    & DANN & 96.6 &  {68.3} & 82.1 & 79.8 & 81.7 \\
    & CDANN & 95.2 & 66.7 &  {85.2} & 80.3 & 81.9 \\
    & CORAL & 96.6 & 67.4 & 84.3 & 81.5 & 82.5 \\
    & MMD & 95.4 & 67.3 & 83.2 & 81.6 & 81.9 \\
    & IIB &  {97.6} & 65.7 & 84.0 & 82.7 & 82.5 \\
    \cmidrule(r){2-7}
    & SAM & 96.6 & 68.1 & 84.9 &  {84.5} &  {83.5} \\
    & SFT (Ours) &  {96.8} &  {68.8} & 84.8 &  {85.8} &  {84.1} \\
    \midrule
    
    \multirow{3}{*}{ViT-L/14} & ERM & 95.8 & 66.8 &  {86.7} & 82.4 & 82.9 \\
    & SAM &  {96.8} &  {68.6} &  {85.5} &  {86.0} &  {84.2} \\
    & SFT (Ours) &  {96.6} &  {68.7} & 85.0 &  {87.2} &  {84.4} \\
    \bottomrule
  \end{tabular}}
\end{table}

\subsection{PACS}
\setlength{\tabcolsep}{18pt}
\begin{table}[H]
  \caption{\small {Out-of-domain accuracies (\%) on each domain of PACS and their average.}}
  \centering
  \scalebox{0.5}{\begin{tabular}{c|l|cccc|c}
    \toprule
     {Backbone} &  {Algorithms} &  {A} &  {C} &  {P} &  {S} &  {Avg}. \\
    \midrule
    \multirow{9}{*}{ViT-B/16} & ERM & 97.7 & 97.5 & 99.6 &  {91.4} &  {96.6} \\
    & IRM &  {98.2} & 96.7 &  {99.8} &  {90.7} & 96.4 \\
    & DANN & 96.6 &  {98.1} & 99.6 & 87.5 & 95.5\\
    & CDANN & 97.4 & 97.8 &  {99.9} & 88.7 & 96.0 \\
    & CORAL & 96.2 & 97.7 & 99.6 & 88.2 & 95.4 \\
    & MMD & 96.9 & 97.7 & 99.6 & 86.0 & 95.1 \\
    & IIB & 97.5 & 97.4 &  {99.8} & 89.2 & 96.0 \\
    \cmidrule(r){2-7}
    & SAM &  {97.8} &  {98.1} & 99.7 & 88.9 & 96.1 \\
    & SFT (Ours) &  {98.2} &  {98.8} &  {99.9} & 90.2 &  {96.8} \\
    \midrule
    
    \multirow{3}{*}{ViT-L/14} & ERM &  {99.2} &  {99.5} &  {99.9} &  {96.6} &  {98.8} \\
    & SAM &  {99.3} &  {99.6} &  {99.9} & 96.0 &  {98.7} \\
    & SFT (Ours) &  {99.3} & 99.3 &  {99.6} &  {96.2} & 98.6 \\
    \bottomrule
  \end{tabular}}
\end{table}
 
\subsection{OfficeHome}
\setlength{\tabcolsep}{18pt}
\begin{table}[H]
  \caption{\small {Out-of-domain accuracies (\%) on each domain of OfficeHome and their average.}}
  \centering
  \scalebox{0.5}{\begin{tabular}{c|l|cccc|c}
    \toprule
     {Backbone} &  {Algorithms} &  {A} &  {C} &  {P} &  {R} &  {Avg}. \\
    \midrule
    \multirow{9}{*}{ViT-B/16} 
    & ERM & 83.7 & 73.8 & 89.9 & 89.2 & 84.1 \\
    & IRM & 81.4 & 73.2 & 88.9 & 88.9 & 83.1 \\
    & DANN & 80.7 & 73.1 & 88.9 & 88.2 & 82.7 \\
    & CDANN & 82.6 & 71.1 & 87.9 & 87.5 & 82.3 \\
    & CORAL & 82.7 & 72.9 & 88.4 & 89.3 & 83.3 \\
    & MMD & 83.5 & 73.0 & 89.5 & 88.6 & 83.7 \\
    & IIB & 81.9 & 73.5 &  {90.7} & 89.5 & 83.9 \\
    & SWAD & 84.9 & 74.9  & 90.8 & 89.9 & 85.1 \\
    \cmidrule(r){2-7}
    & SAM & 84.2 & 76.6 & 91.0 & 90.8 & 85.7 \\
    & SFT (Ours) &  {86.3} &  {77.7} &  {91.1} &  {90.9} &  {86.5} \\
    \midrule
    
    \multirow{3}{*}{ViT-L/14} & ERM & 89.8 & 83.4 & 93.9 & 93.8 & 90.2 \\
    & SAM &  89.8 &  84.7 &  {94.6} &  94.0 &  90.8 \\
    & SFT (Ours) &  {90.8} &  {85.2} &  94.5 &  {94.5} &  {91.3} \\
    \bottomrule
  \end{tabular}}
\end{table}

\subsection{TerraIncognita}
\setlength{\tabcolsep}{18pt}
\begin{table}[H]
  \caption{\small {Out-of-domain accuracies (\%) on each domain of TerraIncognita and their average.}}
  \centering
  \scalebox{0.5}{\begin{tabular}{c|l|cccc|c}
    \toprule
     {Backbone} &  {Algorithms} &  {L100} &  {L38} &  {L43} &  {L46} &  {Avg}. \\
    \midrule
    \multirow{9}{*}{ViT-B/16} & ERM & 58.5 &  {58.2} &  {64.1} & 41.1 & 55.5 \\
    & IRM & 45.7 & 53.5 & 55.4 & 48.8 & 50.9 \\
    & DANN & 52.9 & 52.4 & 56.7 & 45.9 & 52.0 \\
    & CDANN & 58.6 & 51.9 & 61.5 & 47.6 & 54.9 \\
    & CORAL & 51.4 & 45.2 & 60.9 &  {50.6} & 52.0 \\
    & MMD & 57.5 &  {57.1} & 62.0 &  {50.9} & 56.9 \\
    & IIB &  {65.3} & 53.6 &  {65.6} & 47.5 &  {58.0} \\
    \cmidrule(r){2-7}
    & SAM &  {64.6} & 52.0 & 61.5 & 48.3 & 56.6 \\
    & SFT (Ours) & 70.7 & 58.3 & 65.3 &  {50.5} &  {61.2} \\
    \midrule
    
    \multirow{3}{*}{ViT-L/14} & ERM & 65.1 & 55.1 & 69.6 &  {55.4} & 61.3 \\
    & SAM &  {67.3} &  {56.4} &  {72.8} & 54.8 &  {62.8} \\
    & SFT (Ours) &  {65.4} &  {61.1} & 
     {71.2} &  {62.9} &  {65.2} \\
    \bottomrule
  \end{tabular}}
\end{table}

\subsection{DomainNet}
\setlength{\tabcolsep}{12.5pt}
\begin{table}[H]
  \caption{\small {Out-of-domain accuracies (\%) on each domain of TerraIncognita and their average.}}
  \centering
  \scalebox{0.5}{\begin{tabular}{c|l|cccccc|c}
    \toprule
    
     {Backbone} &  {Algorithms} &   {clip}        &  {info}        &  {paint}       &  {quick}       &  {real}        &  {sketch}  &  {Avg.} \\
    \midrule
    \multirow{9}{*}{ViT-B/16} 
    & ERM            & 77.6 & 44.4 & 66.4 & 18.8 & 81.2 & 66.7 & 59.2 \\
    & IRM            & 73.1 & 45.6 & 67.1 & 19.3 & 81.2 & 68.6 & 59.1 \\
    & DANN           & 74.9 & 42.9 & 67.9 & 19.1 & 79.2 & 67.3 & 58.6 \\
    & CDANN          & 74.7 & 44.5 & 66.1 & 19.2 & 79.2 & 67.0 & 58.4 \\
    & CORAL          & 77.6 & 44.7 & 66.6 & 19.1 & 81.0 & 68.1 & 59.5 \\
    & MMD            & 76.8 & 45.9 & 67.4 &  {20.1} & 80.9 & 68.3 & 59.9 \\
    & IIB            & 76.5 & 42.4 & 66.5 & 18.5 & 79.9 & 67.6 & 58.6 \\
    \cmidrule(r){2-9}
    & SAM            & 76.8 & 45.4 & 68.4 & 18.8 & 81.1 & 68.3 & 59.8 \\
    & SFT (Ours) &  {77.9} &  {46.3} &  {68.6} & 19.4 &  {81.6} &  {68.9} &  {60.5} \\
    \midrule
    \multirow{3}{*}{ViT-L/14}
    & ERM             & 82.7 & 54.5 & 73.1 & 23.7 & 84.3 & 74.4 & 65.4 \\
    & SAM\_VPT        & 83.1 & 51.9 & 73.3 & 23.1 & 85.3 & 74.7 & 65.2 \\
    & SFT (Ours)  &  {83.2} &  {55.0} &  {75.2} &  {24.6} &  {86.1} &  {75.3} &  {66.5} \\
    \bottomrule
  \end{tabular}}
\end{table}

\section{Reproducibility}
To guarantee reproducibility, we will provide an explanation of the code and hyperparameters in this section. 
\subsection{Code}\label{section:5}
Our work is built upon DomainBed, which is released under the MIT license. All experiments are conducted on a single NVIDIA Tesla V100 or A40. 

\subsection{Hyperparameters}\label{section:6}
\subsubsection{Experiments on Toy Dataset}
In our toy experiments, we generate a dataset with three classes ($C_1$, $C_2$ and $C_3$) and four domains ($D_1$, $D_2$, $D_3$ and $D_4$). There are $3\times100$ samples in each domain.
We use data from the first three domains ($D_1$, $D_2$ and $D_3$) as the training domains, with the remaining domain $D_4$ serves as the test domain.
For simplicity, we consider the covariance matrices to be diagonal with the same  elements: $\boldsymbol{\Sigma}_{i} = \sigma_{i}^2 \mathbf{I}$ and $\boldsymbol{\Sigma}_{ij} = \sigma_{ij}^2 \mathbf{I}$. 
The specific parameters for each domain and class are provided in Table \ref{Table:params}. During training, we use a linear classifier with the Adam optimizer.
The batch size is set to 16 and the learning rate is 5e-4. 
\setlength{\tabcolsep}{8pt}
\begin{table}[H]
\centering
    \caption{ {Parameters for the generation of the toy dataset.}}\label{Table:params}
    \scalebox{0.8}{
  \begin{tabular}{cccccc}
    \toprule
    Classes & $\mu_{i}$  &  $\sigma_i$ & Domains & $\mu_{ij}$ & $\sigma_{ij}$ \\
    \midrule
    \multirow{4}{*}{$C_1$} & \multirow{4}{*}{$(0,\sqrt{3}/2)$} & \multirow{4}{*}{0.4} & $D_1$ & (0.71,1.03) & 0.2 \\
       
    ~ & ~ & ~& $D_2$ & (-0.04,0.20) & 0.2 \\
       
    ~ & ~ & ~& $D_3$ & (0.08,1.22) & 0.2 \\
       
    ~ & ~ & ~& $D_4$ & (-0.52,0.54) & 0.2 \\
    \midrule

    \multirow{4}{*}{$C_2$} & \multirow{4}{*}{$(-1/2,0)$} & \multirow{4}{*}{0.4} & $D_1$ & (-0.11,0.90) & 0.2 \\
       
    ~ & ~ & ~ & $D_2$ & (-0.45,0.15) & 0.2 \\
       
    ~ & ~ & ~ & $D_3$ & (-0.68,0.03) & 0.2 \\
       
    ~ & ~ & ~ & $D_4$ & (-0.81,-0.11) & 0.2 \\
    \midrule
    \multirow{4}{*}{$C_3$} & \multirow{4}{*}{$(1/2,0)$} & \multirow{4}{*}{0.4} & $D_1$ & (1.25,-0.39) & 0.2 \\
       
    ~ & ~ & ~& $D_2$ & (-0.20,0.52) & 0.2 \\
       
    ~ & ~ & ~& $D_3$ & (0.80,0.23) & 0.2 \\
       
    ~ & ~ & ~& $D_4$ & (0.83,-0.12) & 0.2 \\
    \bottomrule
  \end{tabular}}
\end{table}

\subsubsection{Experiments on Real Dataset}
The hyperparameter search spaces for ResNet-50, ViT-B/16 and ViT-L/14 are shown below. 
In the table, 
$U$ and list indicate Uniform distribution and random choice, respectively.

\setlength{\tabcolsep}{10pt}
\begin{table}[H]
  \caption{\small {Hyperparameter search space for ResNet-50, ViT-B/16 and ViT-L/14.} }
  \centering
  \scalebox{0.6}{\begin{tabular}{l|ccccc}
    \toprule
     {Parameter}   &  {ResNet50}        &  {ViT-B/16}        &  {ViT-L/14}                       \\
    \midrule
    batch size & $U[24,32]$ &  $U[16,28]$ & $U[8,16]$ \\
     
    learning rate & $U[5.0e-6,5.5e-5]$ & $10^{U[-4.5,-3.0]}$ & $10^{U[-4.5,-3.0]}$\\
     
    ResNet dropout & [0.0,0.1,0.5]  & $-$ & $-$\\
     
    weight decay & [1e-4, 1e-6] & 0.0 & 0.0\\
     
    $\rho$ & [0.01, 0.02, 0.03, 0.05, 0.1] & [0.1, 0.2, 0.3, 0.5] & [0.05,0.1,0.2,0.3,0.5] \\
     
    $\lambda_{1}$ & $U[0,1]$ & $U[0,0.5]$ & $U[0,0.5]$ \\
     
    $\lambda_2$ & $U[0,1]$ &  $U[0,0.5]$ & $U[0,0.5]$ \\
     
    $\alpha$ & $10^{U[0.5,3]}$ & $10^{U[0.5,3]}$ & $10^{U[0.5,3]}$ \\
    \bottomrule
  \end{tabular}}
\end{table}

\section{Broader Impacts}\label{section:7}

This paper primarily focuses on developing an effective domain generalization method to address the problem of domain shifts. Given that domain shifts are ubiquitous in real-world applications, this work has the potential to make a positive impact by learning models that are less biased towards ethical aspects. We do not foresee any significant negative social impact of this work.


\begin{thebibliography}{73}
\small
\bibitem[Arjovsky et~al.(2019)Arjovsky, Bottou, Gulrajani, and Lopez-Paz]{IRM}
Martin Arjovsky, L{\'e}on Bottou, Ishaan Gulrajani, and David Lopez-Paz.
\newblock Invariant risk minimization.
\newblock \emph{arXiv preprint arXiv:1907.02893}, 2019.

\bibitem[Bahri et~al.(2021)Bahri, Mobahi, and Tay]{NLPSAM}
Dara Bahri, Hossein Mobahi, and Yi Tay.
\newblock Sharpness-aware minimization improves language model generalization.
\newblock \emph{arXiv preprint arXiv:2110.08529}, 2021.

\bibitem[Bai et~al.(2021{\natexlab{a}})Bai, Sun, Hong, Zhou, Ye, Ye, Chan, and Li]{DecAug}
Haoyue Bai, Rui Sun, Lanqing Hong, Fengwei Zhou, Nanyang Ye, Han-Jia Ye, S-H~Gary Chan, and Zhenguo Li.
\newblock Decaug: Out-of-distribution generalization via decomposed feature representation and semantic augmentation.
\newblock In \emph{Proceedings of the AAAI Conference on Artificial Intelligence}, pages 6705--6713, 2021{\natexlab{a}}.

\bibitem[Bai et~al.(2021{\natexlab{b}})Bai, Zhou, Hong, Ye, Chan, and Li]{Nas-ood}
Haoyue Bai, Fengwei Zhou, Lanqing Hong, Nanyang Ye, S-H~Gary Chan, and Zhenguo Li.
\newblock Nas-ood: Neural architecture search for out-of-distribution generalization.
\newblock In \emph{Proceedings of the IEEE/CVF International Conference on Computer Vision}, pages 8320--8329, 2021{\natexlab{b}}.

\bibitem[Balaji et~al.(2018)Balaji, Sankaranarayanan, and Chellappa]{MetaReg}
Yogesh Balaji, Swami Sankaranarayanan, and Rama Chellappa.
\newblock Metareg: Towards domain generalization using meta-regularization.
\newblock \emph{Advances in neural information processing systems}, 31, 2018.

\bibitem[Beery et~al.(2018)Beery, Van~Horn, and Perona]{TI}
Sara Beery, Grant Van~Horn, and Pietro Perona.
\newblock Recognition in terra incognita.
\newblock In \emph{Proceedings of the European conference on computer vision (ECCV)}, pages 456--473, 2018.

\bibitem[Blanchard et~al.(2021)Blanchard, Deshmukh, Dogan, Lee, and Scott]{MTL}
Gilles Blanchard, Aniket~Anand Deshmukh, {\"U}run Dogan, Gyemin Lee, and Clayton Scott.
\newblock Domain generalization by marginal transfer learning.
\newblock \emph{The Journal of Machine Learning Research}, 22\penalty0 (1):\penalty0 46--100, 2021.

\bibitem[Bui et~al.(2021)Bui, Tran, Tran, and Phung]{mDSDI}
Manh-Ha Bui, Toan Tran, Anh Tran, and Dinh Phung.
\newblock Exploiting domain-specific features to enhance domain generalization.
\newblock \emph{Advances in Neural Information Processing Systems}, 34:\penalty0 21189--21201, 2021.

\bibitem[Cha et~al.(2021)Cha, Chun, Lee, Cho, Park, Lee, and Park]{SWAD}
Junbum Cha, Sanghyuk Chun, Kyungjae Lee, Han-Cheol Cho, Seunghyun Park, Yunsung Lee, and Sungrae Park.
\newblock Swad: Domain generalization by seeking flat minima.
\newblock \emph{Advances in Neural Information Processing Systems}, 34:\penalty0 22405--22418, 2021.

\bibitem[Cha et~al.(2022)Cha, Lee, Park, and Chun]{MIRO}
Junbum Cha, Kyungjae Lee, Sungrae Park, and Sanghyuk Chun.
\newblock Domain generalization by mutual-information regularization with pre-trained models.
\newblock \emph{arXiv preprint arXiv:2203.10789}, 2022.

\bibitem[Chaudhari et~al.(2019)Chaudhari, Choromanska, Soatto, LeCun, Baldassi, Borgs, Chayes, Sagun, and Zecchina]{entropySGD}
Pratik Chaudhari, Anna Choromanska, Stefano Soatto, Yann LeCun, Carlo Baldassi, Christian Borgs, Jennifer Chayes, Levent Sagun, and Riccardo Zecchina.
\newblock Entropy-sgd: Biasing gradient descent into wide valleys.
\newblock \emph{Journal of Statistical Mechanics: Theory and Experiment}, 2019\penalty0 (12):\penalty0 124018, 2019.

\bibitem[Du et~al.(2022)Du, Zhou, Feng, Tan, and Zhou]{SAF}
Jiawei Du, Daquan Zhou, Jiashi Feng, Vincent Tan, and Joey~Tianyi Zhou.
\newblock Sharpness-aware training for free.
\newblock \emph{Advances in Neural Information Processing Systems}, 35:\penalty0 23439--23451, 2022.

\bibitem[Du et~al.(2020)Du, Xu, Xiong, Qiu, Zhen, Snoek, and Shao]{MetaVIB}
Yingjun Du, Jun Xu, Huan Xiong, Qiang Qiu, Xiantong Zhen, Cees~GM Snoek, and Ling Shao.
\newblock Learning to learn with variational information bottleneck for domain generalization.
\newblock In \emph{Computer Vision--ECCV 2020: 16th European Conference, Glasgow, UK, August 23--28, 2020, Proceedings, Part X 16}, pages 200--216. Springer, 2020.

\bibitem[Dziugaite and Roy(2017)]{SharpGenelink2}
Gintare~Karolina Dziugaite and Daniel~M Roy.
\newblock Computing nonvacuous generalization bounds for deep (stochastic) neural networks with many more parameters than training data.
\newblock \emph{arXiv preprint arXiv:1703.11008}, 2017.

\bibitem[Fang et~al.(2013)Fang, Xu, and Rockmore]{VLCS}
Chen Fang, Ye Xu, and Daniel~N Rockmore.
\newblock Unbiased metric learning: On the utilization of multiple datasets and web images for softening bias.
\newblock In \emph{Proceedings of the IEEE International Conference on Computer Vision}, pages 1657--1664, 2013.

\bibitem[Foret et~al.(2020)Foret, Kleiner, Mobahi, and Neyshabur]{SAM}
Pierre Foret, Ariel Kleiner, Hossein Mobahi, and Behnam Neyshabur.
\newblock Sharpness-aware minimization for efficiently improving generalization.
\newblock \emph{arXiv preprint arXiv:2010.01412}, 2020.

\bibitem[Ganin et~al.(2016)Ganin, Ustinova, Ajakan, Germain, Larochelle, Laviolette, Marchand, and Lempitsky]{DANN}
Yaroslav Ganin, Evgeniya Ustinova, Hana Ajakan, Pascal Germain, Hugo Larochelle, Fran{\c{c}}ois Laviolette, Mario Marchand, and Victor Lempitsky.
\newblock Domain-adversarial training of neural networks.
\newblock \emph{The journal of machine learning research}, 17\penalty0 (1):\penalty0 2096--2030, 2016.

\bibitem[Ghosal et~al.(2022)Ghosal, Ming, and Li]{ViTOoDTwo}
Soumya~Suvra Ghosal, Yifei Ming, and Yixuan Li.
\newblock Are vision transformers robust to spurious correlations?
\newblock \emph{arXiv preprint arXiv:2203.09125}, 2022.

\bibitem[Gulrajani and Lopez-Paz(2020)]{DomainBed}
Ishaan Gulrajani and David Lopez-Paz.
\newblock In search of lost domain generalization.
\newblock \emph{arXiv preprint arXiv:2007.01434}, 2020.

\bibitem[Hochreiter and Schmidhuber(1994)]{Sharpness1994}
Sepp Hochreiter and J{\"u}rgen Schmidhuber.
\newblock Simplifying neural nets by discovering flat minima.
\newblock \emph{Advances in neural information processing systems}, 7, 1994.

\bibitem[Hochreiter and Schmidhuber(1997)]{flat1997}
Sepp Hochreiter and J{\"u}rgen Schmidhuber.
\newblock Flat minima.
\newblock \emph{Neural computation}, 9\penalty0 (1):\penalty0 1--42, 1997.

\bibitem[Huang et~al.(2020)Huang, Wang, Xing, and Huang]{RSC}
Zeyi Huang, Haohan Wang, Eric~P Xing, and Dong Huang.
\newblock Self-challenging improves cross-domain generalization.
\newblock In \emph{Computer Vision--ECCV 2020: 16th European Conference, Glasgow, UK, August 23--28, 2020, Proceedings, Part II 16}, pages 124--140. Springer, 2020.

\bibitem[Izmailov et~al.(2018)Izmailov, Podoprikhin, Garipov, Vetrov, and Wilson]{SWA}
Pavel Izmailov, Dmitrii Podoprikhin, Timur Garipov, Dmitry Vetrov, and Andrew~Gordon Wilson.
\newblock Averaging weights leads to wider optima and better generalization.
\newblock \emph{arXiv preprint arXiv:1803.05407}, 2018.

\bibitem[Ji et~al.(2024)Ji, Li, Fu, Afghah, Guo, Yuan, and Ma]{SSAM}
Jie Ji, Gen Li, Jingjing Fu, Fatemeh Afghah, Linke Guo, Xiaoyong Yuan, and Xiaolong Ma.
\newblock A single-step, sharpness-aware minimization is all you need to achieve efficient and accurate sparse training.
\newblock \emph{Advances in Neural Information Processing Systems}, 37:\penalty0 44269--44290, 2024.

\bibitem[Jia et~al.(2022)Jia, Tang, Chen, Cardie, Belongie, Hariharan, and Lim]{VPT}
Menglin Jia, Luming Tang, Bor-Chun Chen, Claire Cardie, Serge Belongie, Bharath Hariharan, and Ser-Nam Lim.
\newblock Visual prompt tuning.
\newblock \emph{arXiv preprint arXiv:2203.12119}, 2022.

\bibitem[Jiang et~al.(2023)Jiang, Yang, Zhang, and Kwok]{AESAM}
Weisen Jiang, Hansi Yang, Yu Zhang, and James Kwok.
\newblock An adaptive policy to employ sharpness-aware minimization.
\newblock \emph{arXiv preprint arXiv:2304.14647}, 2023.

\bibitem[Jiang et~al.(2019)Jiang, Neyshabur, Mobahi, Krishnan, and Bengio]{SharpGenelink3}
Yiding Jiang, Behnam Neyshabur, Hossein Mobahi, Dilip Krishnan, and Samy Bengio.
\newblock Fantastic generalization measures and where to find them.
\newblock \emph{arXiv preprint arXiv:1912.02178}, 2019.

\bibitem[Kaddour et~al.(2022)Kaddour, Liu, Silva, and Kusner]{SWA_SAM}
Jean Kaddour, Linqing Liu, Ricardo Silva, and Matt~J Kusner.
\newblock When do flat minima optimizers work?
\newblock \emph{Advances in Neural Information Processing Systems}, 35:\penalty0 16577--16595, 2022.

\bibitem[Kamath et~al.(2019)Kamath, Liu, and Whitaker]{nlpspeech}
Uday Kamath, John Liu, and James Whitaker.
\newblock \emph{Deep learning for NLP and speech recognition}.
\newblock Springer, 2019.

\bibitem[Keskar et~al.(2016)Keskar, Mudigere, Nocedal, Smelyanskiy, and Tang]{SharpGenelink}
Nitish~Shirish Keskar, Dheevatsa Mudigere, Jorge Nocedal, Mikhail Smelyanskiy, and Ping Tak~Peter Tang.
\newblock On large-batch training for deep learning: Generalization gap and sharp minima.
\newblock \emph{arXiv preprint arXiv:1609.04836}, 2016.

\bibitem[Kim et~al.(2021)Kim, Yoo, Park, Kim, and Lee]{SelfReg}
Daehee Kim, Youngjun Yoo, Seunghyun Park, Jinkyu Kim, and Jaekoo Lee.
\newblock Selfreg: Self-supervised contrastive regularization for domain generalization.
\newblock In \emph{Proceedings of the IEEE/CVF International Conference on Computer Vision}, pages 9619--9628, 2021.

\bibitem[Kingma and Ba(2014)]{Adam}
Diederik~P Kingma and Jimmy Ba.
\newblock Adam: A method for stochastic optimization.
\newblock \emph{arXiv preprint arXiv:1412.6980}, 2014.

\bibitem[Krueger et~al.(2021)Krueger, Caballero, Jacobsen, Zhang, Binas, Zhang, Le~Priol, and Courville]{VREx}
David Krueger, Ethan Caballero, Joern-Henrik Jacobsen, Amy Zhang, Jonathan Binas, Dinghuai Zhang, Remi Le~Priol, and Aaron Courville.
\newblock Out-of-distribution generalization via risk extrapolation (rex).
\newblock In \emph{International Conference on Machine Learning}, pages 5815--5826. PMLR, 2021.

\bibitem[Li et~al.(2022{\natexlab{a}})Li, Zhuang, Fan, and Wang]{CSVPT}
Aodi Li, Liansheng Zhuang, Shuo Fan, and Shafei Wang.
\newblock Learning common and specific visual prompts for domain generalization.
\newblock In \emph{Proceedings of the Asian Conference on Computer Vision}, pages 4260--4275, 2022{\natexlab{a}}.

\bibitem[Li et~al.(2022{\natexlab{b}})Li, Shen, Wang, Zhu, Li, Keutzer, and Zhao]{IIB}
Bo Li, Yifei Shen, Yezhen Wang, Wenzhen Zhu, Dongsheng Li, Kurt Keutzer, and Han Zhao.
\newblock Invariant information bottleneck for domain generalization.
\newblock In \emph{Proceedings of the AAAI Conference on Artificial Intelligence}, pages 7399--7407, 2022{\natexlab{b}}.

\bibitem[Li et~al.(2017)Li, Yang, Song, and Hospedales]{PACS}
Da Li, Yongxin Yang, Yi-Zhe Song, and Timothy~M Hospedales.
\newblock Deeper, broader and artier domain generalization.
\newblock In \emph{Proceedings of the IEEE international conference on computer vision}, pages 5542--5550, 2017.

\bibitem[Li et~al.(2018{\natexlab{a}})Li, Yang, Song, and Hospedales]{MLDG}
Da Li, Yongxin Yang, Yi-Zhe Song, and Timothy Hospedales.
\newblock Learning to generalize: Meta-learning for domain generalization.
\newblock In \emph{Proceedings of the AAAI conference on artificial intelligence}, 2018{\natexlab{a}}.

\bibitem[Li et~al.(2018{\natexlab{b}})Li, Pan, Wang, and Kot]{MMD}
Haoliang Li, Sinno~Jialin Pan, Shiqi Wang, and Alex~C Kot.
\newblock Domain generalization with adversarial feature learning.
\newblock In \emph{Proceedings of the IEEE conference on computer vision and pattern recognition}, pages 5400--5409, 2018{\natexlab{b}}.

\bibitem[Li et~al.(2018{\natexlab{c}})Li, Xu, Taylor, Studer, and Goldstein]{LossVisual}
Hao Li, Zheng Xu, Gavin Taylor, Christoph Studer, and Tom Goldstein.
\newblock Visualizing the loss landscape of neural nets.
\newblock \emph{Advances in neural information processing systems}, 31, 2018{\natexlab{c}}.

\bibitem[Li et~al.(2018{\natexlab{d}})Li, Gong, Tian, Liu, and Tao]{cDANN}
Ya Li, Mingming Gong, Xinmei Tian, Tongliang Liu, and Dacheng Tao.
\newblock Domain generalization via conditional invariant representations.
\newblock In \emph{Proceedings of the AAAI conference on artificial intelligence}, 2018{\natexlab{d}}.

\bibitem[Li et~al.(2020)Li, Yang, Wang, and Xu]{AdaptNAS}
Yanxi Li, Zhaohui Yang, Yunhe Wang, and Chang Xu.
\newblock Adapting neural architectures between domains.
\newblock \emph{Advances in Neural Information Processing Systems}, 33:\penalty0 789--798, 2020.

\bibitem[Liang et~al.(2024)Liang, Song, Zheng, Wang, Yu, Li, Li, Xiong, and Li]{selfLLM}
Xun Liang, Shichao Song, Zifan Zheng, Hanyu Wang, Qingchen Yu, Xunkai Li, Rong-Hua Li, Feiyu Xiong, and Zhiyu Li.
\newblock Internal consistency and self-feedback in large language models: A survey.
\newblock \emph{arXiv preprint arXiv:2407.14507}, 2024.

\bibitem[Lopez and Kalita(2017)]{NLP}
Marc~Moreno Lopez and Jugal Kalita.
\newblock Deep learning applied to nlp.
\newblock \emph{arXiv preprint arXiv:1703.03091}, 2017.

\bibitem[Madaan et~al.(2024)Madaan, Tandon, Gupta, Hallinan, Gao, Wiegreffe, Alon, Dziri, Prabhumoye, Yang, et~al.]{selfrefineLLM}
Aman Madaan, Niket Tandon, Prakhar Gupta, Skyler Hallinan, Luyu Gao, Sarah Wiegreffe, Uri Alon, Nouha Dziri, Shrimai Prabhumoye, Yiming Yang, et~al.
\newblock Self-refine: Iterative refinement with self-feedback.
\newblock \emph{Advances in Neural Information Processing Systems}, 36, 2024.

\bibitem[Mobahi(2016)]{sharpdiffusion}
Hossein Mobahi.
\newblock Training recurrent neural networks by diffusion.
\newblock \emph{arXiv preprint arXiv:1601.04114}, 2016.

\bibitem[M{\"u}ller et~al.(2019)M{\"u}ller, Kornblith, and Hinton]{LabelSmooth}
Rafael M{\"u}ller, Simon Kornblith, and Geoffrey~E Hinton.
\newblock When does label smoothing help?
\newblock \emph{Advances in neural information processing systems}, 32, 2019.

\bibitem[Nam et~al.(2019)Nam, Lee, Park, Yoon, and Yoo]{Sagnet}
Hyeonseob Nam, HyunJae Lee, Jongchan Park, Wonjun Yoon, and Donggeun Yoo.
\newblock Reducing domain gap via style-agnostic networks.
\newblock \emph{arXiv preprint arXiv:1910.11645}, 2\penalty0 (7):\penalty0 8, 2019.

\bibitem[Parascandolo et~al.(2020)Parascandolo, Neitz, Orvieto, Gresele, and Sch{\"o}lkopf]{AndMask}
Giambattista Parascandolo, Alexander Neitz, Antonio Orvieto, Luigi Gresele, and Bernhard Sch{\"o}lkopf.
\newblock Learning explanations that are hard to vary.
\newblock \emph{arXiv preprint arXiv:2009.00329}, 2020.

\bibitem[Peng et~al.(2019)Peng, Bai, Xia, Huang, Saenko, and Wang]{DN}
Xingchao Peng, Qinxun Bai, Xide Xia, Zijun Huang, Kate Saenko, and Bo Wang.
\newblock Moment matching for multi-source domain adaptation.
\newblock In \emph{Proceedings of the IEEE/CVF international conference on computer vision}, pages 1406--1415, 2019.

\bibitem[Radford et~al.(2021)Radford, Kim, Hallacy, Ramesh, Goh, Agarwal, Sastry, Askell, Mishkin, Clark, et~al.]{CLIP}
Alec Radford, Jong~Wook Kim, Chris Hallacy, Aditya Ramesh, Gabriel Goh, Sandhini Agarwal, Girish Sastry, Amanda Askell, Pamela Mishkin, Jack Clark, et~al.
\newblock Learning transferable visual models from natural language supervision.
\newblock In \emph{International Conference on Machine Learning}, pages 8748--8763. PMLR, 2021.

\bibitem[Sagawa et~al.(2019)Sagawa, Koh, Hashimoto, and Liang]{DRO}
Shiori Sagawa, Pang~Wei Koh, Tatsunori~B Hashimoto, and Percy Liang.
\newblock Distributionally robust neural networks for group shifts: On the importance of regularization for worst-case generalization.
\newblock \emph{arXiv preprint arXiv:1911.08731}, 2019.

\bibitem[Shi et~al.(2021)Shi, Seely, Torr, Siddharth, Hannun, Usunier, and Synnaeve]{Fish}
Yuge Shi, Jeffrey Seely, Philip~HS Torr, N Siddharth, Awni Hannun, Nicolas Usunier, and Gabriel Synnaeve.
\newblock Gradient matching for domain generalization.
\newblock \emph{arXiv preprint arXiv:2104.09937}, 2021.

\bibitem[Shim et~al.(2023)Shim, Jung, and Kinnunen]{SpeechSAM}
Hye-jin Shim, Jee-weon Jung, and Tomi Kinnunen.
\newblock Multi-dataset co-training with sharpness-aware optimization for audio anti-spoofing.
\newblock \emph{arXiv preprint arXiv:2305.19953}, 2023.

\bibitem[Sun and Saenko(2016)]{CORAL}
Baochen Sun and Kate Saenko.
\newblock Deep coral: Correlation alignment for deep domain adaptation.
\newblock In \emph{European conference on computer vision}, pages 443--450. Springer, 2016.

\bibitem[Vapnik(1998)]{ERM}
Vladimir Vapnik.
\newblock Statistical learning theory.
\newblock \emph{(No Title)}, 1998.

\bibitem[Venkateswara et~al.(2017)Venkateswara, Eusebio, Chakraborty, and Panchanathan]{OH}
Hemanth Venkateswara, Jose Eusebio, Shayok Chakraborty, and Sethuraman Panchanathan.
\newblock Deep hashing network for unsupervised domain adaptation.
\newblock In \emph{Proceedings of the IEEE conference on computer vision and pattern recognition}, pages 5018--5027, 2017.

\bibitem[Volpi et~al.(2018)Volpi, Namkoong, Sener, Duchi, Murino, and Savarese]{ADA}
Riccardo Volpi, Hongseok Namkoong, Ozan Sener, John~C Duchi, Vittorio Murino, and Silvio Savarese.
\newblock Generalizing to unseen domains via adversarial data augmentation.
\newblock \emph{Advances in neural information processing systems}, 31, 2018.

\bibitem[Voulodimos et~al.(2018)Voulodimos, Doulamis, Doulamis, and Protopapadakis]{CV}
Athanasios Voulodimos, Nikolaos Doulamis, Anastasios Doulamis, and Eftychios Protopapadakis.
\newblock Deep learning for computer vision: A brief review.
\newblock \emph{Computational intelligence and neuroscience}, 2018, 2018.

\bibitem[Wang et~al.(2022)Wang, Lan, Liu, Ouyang, Qin, Lu, Chen, Zeng, and Yu]{DGSurvey}
Jindong Wang, Cuiling Lan, Chang Liu, Yidong Ouyang, Tao Qin, Wang Lu, Yiqiang Chen, Wenjun Zeng, and Philip Yu.
\newblock Generalizing to unseen domains: A survey on domain generalization.
\newblock \emph{IEEE Transactions on Knowledge and Data Engineering}, 2022.

\bibitem[Wang and Deng(2018)]{DA}
Mei Wang and Weihong Deng.
\newblock Deep visual domain adaptation: A survey.
\newblock \emph{Neurocomputing}, 312:\penalty0 135--153, 2018.

\bibitem[Wang et~al.(2023)Wang, Zhang, Lei, and Zhang]{SAGM}
Pengfei Wang, Zhaoxiang Zhang, Zhen Lei, and Lei Zhang.
\newblock Sharpness-aware gradient matching for domain generalization.
\newblock In \emph{Proceedings of the IEEE/CVF Conference on Computer Vision and Pattern Recognition}, pages 3769--3778, 2023.

\bibitem[White et~al.(2023)White, Safari, Sukthanker, Ru, Elsken, Zela, Dey, and Hutter]{NAS1000}
Colin White, Mahmoud Safari, Rhea Sukthanker, Binxin Ru, Thomas Elsken, Arber Zela, Debadeepta Dey, and Frank Hutter.
\newblock Neural architecture search: Insights from 1000 papers.
\newblock \emph{arXiv preprint arXiv:2301.08727}, 2023.

\bibitem[Zhang et~al.(2022)Zhang, Zhang, Zhang, Jin, Zhou, Cai, Zhao, Liu, and Liu]{ViTOoD}
Chongzhi Zhang, Mingyuan Zhang, Shanghang Zhang, Daisheng Jin, Qiang Zhou, Zhongang Cai, Haiyu Zhao, Xianglong Liu, and Ziwei Liu.
\newblock Delving deep into the generalization of vision transformers under distribution shifts.
\newblock In \emph{Proceedings of the IEEE/CVF Conference on Computer Vision and Pattern Recognition}, pages 7277--7286, 2022.

\bibitem[Zhang et~al.(2017)Zhang, Cisse, Dauphin, and Lopez-Paz]{Mixup}
Hongyi Zhang, Moustapha Cisse, Yann~N Dauphin, and David Lopez-Paz.
\newblock mixup: Beyond empirical risk minimization.
\newblock \emph{arXiv preprint arXiv:1710.09412}, 2017.

\bibitem[Zhang et~al.(2020)Zhang, Marklund, Gupta, Levine, and Finn]{ARM}
Marvin Zhang, Henrik Marklund, Abhishek Gupta, Sergey Levine, and Chelsea Finn.
\newblock Adaptive risk minimization: A meta-learning approach for tackling group shift.
\newblock \emph{arXiv preprint arXiv:2007.02931}, 8:\penalty0 9, 2020.

\bibitem[Zhang et~al.(2021)Zhang, Cui, Xu, Zhou, He, and Shen]{stable}
Xingxuan Zhang, Peng Cui, Renzhe Xu, Linjun Zhou, Yue He, and Zheyan Shen.
\newblock Deep stable learning for out-of-distribution generalization.
\newblock In \emph{Proceedings of the IEEE/CVF Conference on Computer Vision and Pattern Recognition}, pages 5372--5382, 2021.

\bibitem[Zhang et~al.(2023{\natexlab{a}})Zhang, Xu, Yu, Dong, Tian, and Cui]{FAD}
Xingxuan Zhang, Renzhe Xu, Han Yu, Yancheng Dong, Pengfei Tian, and Peng Cui.
\newblock Flatness-aware minimization for domain generalization.
\newblock In \emph{Proceedings of the IEEE/CVF International Conference on Computer Vision}, pages 5189--5202, 2023{\natexlab{a}}.

\bibitem[Zhang et~al.(2023{\natexlab{b}})Zhang, Xu, Yu, Zou, and Cui]{GAM}
Xingxuan Zhang, Renzhe Xu, Han Yu, Hao Zou, and Peng Cui.
\newblock Gradient norm aware minimization seeks first-order flatness and improves generalization.
\newblock In \emph{Proceedings of the IEEE/CVF Conference on Computer Vision and Pattern Recognition}, pages 20247--20257, 2023{\natexlab{b}}.

\bibitem[Zhang et~al.(2018)Zhang, Geiger, Pohjalainen, Mousa, Jin, and Schuller]{speech}
Zixing Zhang, J{\"u}rgen Geiger, Jouni Pohjalainen, Amr El-Desoky Mousa, Wenyu Jin, and Bj{\"o}rn Schuller.
\newblock Deep learning for environmentally robust speech recognition: An overview of recent developments.
\newblock \emph{ACM Transactions on Intelligent Systems and Technology (TIST)}, 9\penalty0 (5):\penalty0 1--28, 2018.

\bibitem[Zheng et~al.(2022)Zheng, Yue, Wang, and You]{DoPrompt}
Zangwei Zheng, Xiangyu Yue, Kai Wang, and Yang You.
\newblock Prompt vision transformer for domain generalization.
\newblock \emph{arXiv preprint arXiv:2208.08914}, 2022.

\bibitem[Zhou et~al.(2021)Zhou, Yang, Qiao, and Xiang]{Mixstyle}
Kaiyang Zhou, Yongxin Yang, Yu Qiao, and Tao Xiang.
\newblock Domain generalization with mixstyle.
\newblock \emph{arXiv preprint arXiv:2104.02008}, 2021.

\bibitem[Zhuang et~al.(2022)Zhuang, Gong, Yuan, Cui, Adam, Dvornek, Tatikonda, Duncan, and Liu]{GSAM}
Juntang Zhuang, Boqing Gong, Liangzhe Yuan, Yin Cui, Hartwig Adam, Nicha Dvornek, Sekhar Tatikonda, James Duncan, and Ting Liu.
\newblock Surrogate gap minimization improves sharpness-aware training.
\newblock \emph{arXiv preprint arXiv:2203.08065}, 2022.

\bibitem[Zou et~al.(2024)Zou, Kawaguchi, Liu, Liu, Lee, and Hsu]{SAMDG2024}
Yingtian Zou, Kenji Kawaguchi, Yingnan Liu, Jiashuo Liu, Mong-Li Lee, and Wynne Hsu.
\newblock Towards robust out-of-distribution generalization bounds via sharpness.
\newblock \emph{arXiv preprint arXiv:2403.06392}, 2024.

\bibitem[McAllester(1999)]{McAllester}
David~A McAllester.
\newblock Pac-bayesian model averaging.
\newblock In \emph{Proceedings of the twelfth annual conference on Computational learning theory}, pages 164--170, 1999.

\end{thebibliography}
\end{document}